# Flaw Selection Strategies For Partial-Order Planning

**Martha E. Pollack**                                    POLLACK@CS.PITT.EDU
*Department of Computer Science and Intelligent Systems Program,*
*University of Pittsburgh, Pittsburgh, PA 15260 USA*

**David Joslin**                                         JOSLIN@CIRL.UOREGON.EDU
*Computational Intelligence Research Laboratory,*
*University of Oregon, Eugene, OR 97403 USA*

**Massimo Paolucci**                                     PAOLUCCI@PITT.EDU
*Intelligent Systems Program,*
*University of Pittsburgh, Pittsburgh, PA 15260 USA*

## Abstract

Several recent studies have compared the relative efficiency of alternative flaw selection strategies for partial-order causal link (POCL) planning. We review this literature, and present new experimental results that generalize the earlier work and explain some of the discrepancies in it. In particular, we describe the Least-Cost Flaw Repair (LCFR) strategy developed and analyzed by Joslin and Pollack (1994), and compare it with other strategies, including Gerevini and Schubert's (1996) ZLIFO strategy. LCFR and ZLIFO make very different, and apparently conflicting claims about the most effective way to reduce search-space size in POCL planning. We resolve this conflict, arguing that much of the benefit that Gerevini and Schubert ascribe to the LIFO component of their ZLIFO strategy is better attributed to other causes. We show that for many problems, a strategy that combines least-cost flaw selection with the delay of separable threats will be effective in reducing search-space size, and will do so without excessive computational overhead. Although such a strategy thus provides a good default, we also show that certain domain characteristics may reduce its effectiveness.

## 1. Introduction

Much of the current research in plan generation centers on partial-order causal link (POCL) algorithms, which descend from McAllester and Rosenblitt's (1991) SNLP algorithm. POCL planning involves searching through a space of partial plans, where the successors of a node representing partial plan $P$ are refinements of $P$. As with any search problem, POCL planning requires effective search control strategies.

In POCL planning, search control has two components. The first, *node selection*, involves choosing which partial plan to refine next. Once a partial plan has been selected for refinement, the planner must then perform *flaw selection,* which involves choosing either a threat to resolve or an open condition to establish.

Over the past few years, several studies have compared the relative efficiency of alternative flaw selection strategies for POCL planning and their extensions (Peot & Smith, 1993; Joslin & Pollack, 1994; Srinivasan & Howe, 1995; Gerevini & Schubert, 1996; Williamson & Hanks, 1996). These studies have been motivated at least in part by a tension between the attractive formal properties of the POCL algorithms, and the limitations in putting them to practical use that result from their relatively poor performance. To date, the POCL algorithms cannot match the efficiency of the so-called industrial-strength planners such as SIPE (Wilkins, 1988; Wilkins & Desimone, 1994) and O-Plan (Currie & Tate, 1991; Tate, Drabble, & Dalton, 1994). Flaw selection strategy has been shown to have a significant effect on the efficiency of POCL planning algorithms, and thus researchers have viewed the design of improved flaw selection strategies as one means of making POCL planning algorithms more practical.

In this paper, we review the literature on flaw selection strategies, and present new experimental results that generalize the earlier work and explain some of the discrepancies in it. In particular, we describe the Least-Cost Flaw Repair (LCFR) strategy developed and analyzed by Joslin and Pollack (1994), and compare it with other strategies, including Gerevini and Schubert's ZLIFO strategy (1996). LCFR and ZLIFO make very different, and apparently conflicting claims about the most effective way to reduce search-space size in POCL planning. We resolve this conflict, arguing that much of the benefit that Gerevini and Schubert ascribe to the LIFO component of their ZLIFO strategy is better attributed to other causes. We show that for many problems, a strategy that combines least-cost flaw selection with the delay of separable threats will be effective in reducing search-space size, and will do so without excessive computational overhead. Although such a strategy thus provides a good default, we also show that certain domain characteristics may reduce its effectiveness.

## 2. Background

### 2.1 Node and Flaw Selection

Although the main ideas of POCL planning have been in the literature for more than two decades, serious efforts at comparing alternative plan generation algorithms have been relatively recent. What made these comparisons possible was the development of a set of clear algorithms with provable formal properties, notably TWEAK (Chapman, 1987), and SNLP (McAllester & Rosenblitt, 1991). These algorithms were not intended to add functionality to known planning methods, but rather to capture the essential elements of these known methods in a readily analyzable fashion.

In analyzing POCL algorithms, researchers have found it useful to decouple the search control strategy from the underlying plan refinement process. Figure 1 is a generic POCL algorithm, in which we highlight the two search decisions.[1] Following convention, we use **CHOOSE** to indicate that node selection is a backtracking point, and **SELECT** to indicate that flaw selection is not. A given node may not lead to a solution, and so it may be necessary to backtrack and consider alternative nodes. On the other hand, if a node does lead to a solution, that solution will be found regardless of the order in which its flaws are selected. See Weld's (1994) tutorial paper for more discussion of this difference.

The generic algorithm sketched in the figure must be supplemented with search strategies that implement the **CHOOSE** and **SELECT** operators. Most POCL algorithms perform node selection using a best-first ranking that computes some function of the number of

---

1. Various versions of this well-known algorithm have appeared in the literature (Weld, 1994; Russell & Norvig, 1995; Kambhampati, Knoblock, & Yang, 1995). The version we give corresponds most directly to that given by Williamson and Hanks (1996).

**POCL** (*init,goal*)
    *dummy-plan* ← make-skeletal-plan(*init,goal*).
    *nodes* ← { *dummy-plan* }.
    While *nodes* is not empty do:
        **CHOOSE (and remove) a partial plan $P$ from *nodes*. (Node Selection)**
        If $P$ has no flaws
            then return $P$
            else do:
                **SELECT a flaw from $P$. (Flaw Selection)**
                Add all refinements of $P$ to *nodes*.
    Return failure (because *nodes* has become empty without a flaw-free plan being found.)

Figure 1: The Basic POCL Planning Algorithm

steps (denoted $S$), open conditions ($OC$), and unsafe conditions ($UC$, *i.e.*, threats) in the partial plan. Gerevini and Schubert (1996) have argued that, in general, only steps and open conditions should be included in the ranking function, and we adopt that strategy in our experiments, except where otherwise indicated.

Having chosen a node, a POCL planning algorithm must then select a flaw—open condition or threat—within that node to repair. Open conditions are repaired by *establishment*, which consists either in adding a new step that has a unifying condition as an effect (along with a causal link from that new step to the condition), or else in simply adding a new causal link from an existing step with a unifying effect. We use the term *repair cost* to denote the number of possible ways to repair a flaw.

For an open condition $o$, the repair cost $R(o)$ is $I + S + N$, where

    $I$ = the number of conditions in the initial state that unify with $o$ given the current binding constraints,

    $S$ = the number of conditions in the effects of existing plan steps that unify with $o$ given the current binding constraints, counting only existing plan steps that are not constrained to occur after the step associated with $o$, and

    $N$ = the number of conditions in the effects of operators in the library that unify with $o$ given the current binding constraints.

Note that over time, the repair cost for an open condition that is not resolved may either increase, as new steps that might achieve the condition are added to the plan, or decrease, as steps already in the plan are constrained by temporal ordering or variable binding so that they can no longer achieve the condition.

In considering the cost of threat repair, it is useful to distinguish between *nonseparable* and *separable* threats. Nonseparable threats consist of a step $S_1$ with effect $E$, and a causal link $\prec S_2, F, S_3 \succ$, where $E$ and $F$ are complementary literals that necessarily unify: either they are complementary ground literals ($E \equiv \neg F$), or else they are complementary literals where each of $E$'s variables is identical with, or forced by a binding constraint to be equivalent to the variable in the same position in $F$ (e.g., $E = p(x, y)$ and $F = \neg p(x, z)$, where there is currently a binding constraint that $y = z$).[2]

Nonseparable threats can be repaired in at most two ways: by *promoting* $S_1$, requiring it to occur after $S_3$, or by *demoting* it, requiring it to occur before $S_2$. Of course, already existing temporal ordering constraints may block one or both or these repair options, which is why there are *at most* two possible repairs.[3] Over time, the repair cost for an unresolved nonseparable threat can only decrease.

A separable threat consists of a step $S_1$ with an effect $E$, and a causal link $\prec S_2, F, S_3 \succ$, where $E$ and $F$ are complementary literals that *can* be unified, but where such a unification is not forced (e.g., where $E = p(x)$ and $F = p(y)$ and there does not exist a binding constraint $x = y$). In such circumstances, the threat may disappear if a subsequent variable binding blocks the unification. (A nonseparable threat may also disappear if a subsequent ordering constraint has the effect of imposing promotion or demotion.) The repair cost for a separable threat may be higher than that for an nonseparable threat: not only can promotion and demotion be used, but so can *separation*, which involves forcing a variable binding that blocks the unification. Separation can introduce one repair for each unbound variable in the threat. For example, if the effect $P(x, y, z)$ threatens $\prec S_2, \neg P(t, u, v), S_3 \succ$, there are three possible repairs: $x \neq t$, $y \neq u$, and $z \neq v$. As with nonseparable threats, the repair cost for a separable threat that remains unresolved can only decrease over time.

## 2.2 Notation

The flaw selection strategies that have been discussed in the literature typically have been given idiosyncratic names (e.g., DUnf, LCFR, ZLIFO). It is useful, in comparing them, to have a precise unifying notation. We therefore specify a flaw strategy as a sequence of preferences. A strategy begins by attempting to find a flaw that satisfies its first preference; if it is unable to do so, it then looks for a flaw that satisfies the second preference; and so on. To ensure that a POCL algorithm using the strategy is complete, the sequence of preferences must be exhaustive: every flaw must satisfy *some* preference. If a flaw satisfies more than one preference in a strategy, we assume that the first match is what counts.

In principle, a preference could identify any feature of a flaw. In practice, however, flaw selection strategies have only made use of a small number of features: the type of a flaw (open condition, nonseparable threat, or separable threat), the number of ways it can be repaired, and the time at which it was introduced into the plan. Often, more than one flaw will have a given feature, in which case a tie-breaking strategy may be specified for choosing among the relevant flaws.

We therefore describe a preference using the following notation

$$\{\textit{flaw types}\}_{\textit{repair cost range}}\textit{tie-breaking strategy}$$

---

2. An alternative approach also treats cases in which $E \equiv F$ as threats; this is required to make the planner *systematic*, i.e., guaranteed never to generate the same node more than once (McAllester & Rosenblitt, 1991).

3. Conditional planners make use of an additional method of threat resolution—confrontation—but we ignore that within this paper (Peot & Smith, 1992; Etzioni, Hanks, Weld, Draper, Lesh, & Williamson, 1992). Joslin (1996) provides a detailed account of generalizing the treatment of flaws to other types of planning decisions.

which indicates a preference for any flaw $f$ of the specified type or types, provided that the repair cost for $f$ falls within the range of values specified. (If there are no restrictions on repair cost, we omit the *repair cost range*.) If more than one flaw meets these criteria, then the tie-breaking strategy is applied to select among them.

We abbreviate flaw type as "o" (for open condition), "n" (for nonseparable threat), and "s" (for separable threat). We also use abbreviations for common tie-breaking strategies, e.g., "LC" (least (repair) cost), "LIFO" and "R" (Random). In the case of LC, if a choice must be made between flaws that have the same repair cost, LIFO selection is used.

Thus, for example

$$\{n\}_{0\text{-}1}R$$

specifies a preference for nonseparable threats with a repair cost of zero or one; if more than one flaw meets these conditions, a random selection will be made among them. We use the term *forced* to describe flaws with repair cost of one or less.

An example of a complete flaw selection strategy is then:

$$\{n\}_{0\text{-}1}R \ / \ \{o\}LIFO \ / \ \{n,s\}R$$

This strategy would begin by looking for a forced nonseparable threat; if more than one flaw meets this criterion, the strategy would select randomly among them. If there are no forced nonseparable threats, it would then look for an open condition, with any repair cost, using a LIFO scheme to select among them. Finally, if there are neither forced nonseparable threats nor open conditions, it would randomly select either an unforced nonseparable threat or a separable threat.

While we have distinguished between flaw type and maximum repair cost, on the one hand, and tie-breaking strategy, on the other, it is easy to describe strategies that use something other than flaw type as the main criterion for selection. For example, a pure LIFO selection strategy would be encoded as follows. (Henceforth, we give the name of a strategy in boldface preceding the specification.)

$$\textbf{LIFO} \qquad \{o,n,s\}LIFO$$

## 3. Flaw Selection Strategies

We begin by reviewing the flaw selection strategies that have been proposed and studied in the literature to date.

### 3.1 Threat Preference and Delay

The original SNLP algorithm (McAllester & Rosenblitt, 1991) adopted a flaw selection strategy in which threats are resolved before open conditions, and early versions of the widely used UCPOP planning system (Penberthy & Weld, 1992) did the same.[4] SNLP does not specify a principle for selecting among multiple threats or multiple opens; UCPOP used LIFO for this purpose. Employing the notation above, we can describe the basic UCPOP strategy as:

---

4. In the current version of UCPOP (v.4), the flaw selection strategy that is run by default is the DSep strategy, discussed just below. For historical reasons, we maintain the name DSep for that strategy, and use UCPOP for the older default strategy.

**UCPOP**     {n,s}LIFO / {o}LIFO

The first study of alternative flaw selection strategies was done by Peot and Smith (1993), who relaxed the requirement that threats always be resolved before open conditions, and examined several strategies for delaying the resolution of some threats. They analyzed five different strategies for delaying the repair of threats; of these, two are provably superior: DSep and DUnf.

DSep (Delay Separable Threats) was motivated by the observation that sometimes separable threats can simply disappear in the planning process as blocking variable bindings are introduced. As we pointed out earlier, nonseparable threats may also "disappear", but typically this is less frequent. Moreover, if the resolution of *all* threats—separable and nonseparable—were delayed, then nonseparable threats would only disappear early as a side effect of step reuse, making their disappearance even less frequent.

The DSep strategy therefore defers the repair of all separable threats until the very end of the planning process. However, like UCPOP, it continues to give preference to nonseparable threats:

**DSep**     {n}LIFO / {o}LIFO / {s}LIFO

Actually, Peot and Smith do not specify a tie-breaking strategy for choosing among multiple threats; we have here indicated this as LIFO. They explored three different tie-breaking strategies for selecting open conditions (FIFO, LIFO, and least-cost); here we list LIFO, but one can also specify the alternatives:

**DSep-LC**     {n}LIFO / {o}LC / {s}LIFO
**DSep-FIFO**     {n}LIFO / {o}FIFO / {s}LIFO

Peot and Smith prove that the search space generated by a POCL planner using DSep will never be larger than the search space generated using the UCPOP strategy. This result holds when the tie-breaking strategy for open conditions is LIFO or FIFO, but not LC, a point we will return to later in the paper.

Peot and Smith's second successful strategy is DUnf (Delay Unforced Threats). It makes use of the notion of forced flaws. As we stated earlier, a flaw is forced if there is at most one possible way to repair it. The DUnf strategy delays the repair of all unforced threats:

**DUnf**     ${n,s}_0$LIFO / ${n,s}_1$LIFO / {o}LIFO / ${n,s}_{2\text{-}\infty}$LIFO

We can define DUnf-LC and DUnf-FIFO in a manner analogous to that used for DSep-LC and DSep-FIFO:

**DUnf-LC**     ${n,s}_0$LIFO / ${n,s}_1$LIFO / {o}LC / ${n,s}_{2\text{-}\infty}$LIFO
**DUnf-FIFO**     ${n,s}_0$LIFO / ${n,s}_1$LIFO / {o}FIFO / ${n,s}_{2\text{-}\infty}$LIFO

Peot and Smith proved that the DUnf strategy would never generate a larger search space than either of the remaining two strategies that they examined. They also proved that that DSep and DUnf are incomparable: there exist planning problems for which DSep generates a smaller search space than DUnf, and other problems for which the reverse is true.

Peot and Smith support their theoretical results on DSep and DUnf with experiments showing that, at least for the domains they examined, these strategies can result in significant decrease in search-space size. The decrease in search is correlated with the difficulty of the problem, and consequently, as the problems get more difficult, these strategies reduce search time as well as space. That is, on large enough problems, they "pay for" their own overhead.

In follow-on work, Peot and Smith (1994) describe a strategy called DMin, which generates smaller search spaces than both DSep and DUnf. DMin combines a process of pruning dead-end nodes with the process of flaw selection. It gives preference to forced threats. If there are no forced threats, it checks to see whether *all* the remaining nonseparable threats could be repaired simultaneously. If so, it leaves them as threats, and selects an open condition to repair; if there are no open conditions, then presumably it selects a remaining unforced threat to repair. On the other hand, if it is impossible to repair all the unforced, nonseparable threats, then the node is a dead end, and can be pruned from the search space. Note that some dead-end nodes can be recognized immediately, even without doing the complete consistency checking of DMin. This is because an unrepairable flaw cannot subsequently become repairable, hence, any node containing a flaw with repair cost of zero is a dead end. Consequently, all flaw selection strategies should give highest priority to such flaws (Joslin & Pollack, 1996; Joslin, 1996).

## 3.2 Least-Cost Flaw Repair

Peot and Smith's work provided the foundation for our subsequent exploration of the least-cost flaw repair (LCFR) strategy (Joslin & Pollack, 1994). We hypothesized that the power of the DUnf strategy might come not from its relative ordering of threats and open conditions, but instead from the fact that DUnf has the effect of imposing a partial preference for least-cost flaw selection. DUnf will always prefer a forced threat, which, by definition has a repair cost of at most one; thus, in cases in which there is a forced threat, DUnf will make a low-cost selection. What about cases in which there are no forced threats? Then DUnf will have to select among open conditions, assuming there are any. If our hypothesis is correct, a version of DUnf that makes *this* selection using a least-cost strategy (i.e., DUnf-LC) ought to perform better than a version that uses one of the other strategies (i.e., bare DUnf or DUnf-FIFO). In fact, if it is the selection of low-cost repairs that is causing the search-space reduction, then the idea of treating threat resolution differently from open condition establishment ought to be abandoned. Instead, a strategy that always selects the flaw with minimal repair cost, regardless of whether it is a threat or an open condition, ought to show the best performance. This is the Least-Cost Flaw Repair (LCFR) strategy:[5]

$$\textbf{LCFR} \qquad \{\text{o,n,s}\}\text{LC}$$

There are strong similarities between LCFR and certain heuristics that have been proposed and studied in the literature on constraint satisfaction problems (CSPs). This is perhaps not surprising, given that flaw selection in POCL planning corresponds in some fairly strong ways to variable selection in constraint programming. Flaws in a POCL planner represent decisions that are yet to be made, and that must be made before the plan will be complete; unbound variables play a similar role in constraint satisfaction problems (CSPs).[6] Although there exist a number of heuristics for selecting a variable to branch on in solving a CSP (Kumar, 1992), one well-known heuristic that is often quite effective is the *fail first principle*, which picks the variable that is the "most constrained" when selecting a variable to branch on. A simple and common implementation of the fail first principle selects the variable with the smallest domain (Tsang, 1993).

---

5. The LCFR strategy is similar to the branch-1/branch-n search heuristics included in the O-Plan system (Currie & Tate, 1991). The contribution of our original work on this topic was to isolate this strategy and examine it in detail.

The intuition behind the fail first principle is that one should prune dead-end regions of the search as early as possible. The unbound variables that are most tightly constrained are likely to be points at which the current partial solution is most "brittle" in some sense, and by branching on those variables we hope to find a contradiction (if one exists) quickly. Similarly, LCFR can be thought of as selecting the "most constrained" flaws, resulting in better pruning.

A similar heuristic has also been adopted in recent work on controlling search in hierarchical task network (HTN) planning, in the Dynamic Variable Commitment Strategy (DVCS). DVCS, like LCFR, is based on a minimal-branching heuristic. Experimental analyses demonstrate that DVCS generally produces a well-focused search (Tsuneto, Erol, Hendler, & Nau, 1996).

Our own initial experimental results, presented in Joslin and Pollack (1994), similarly supported the hypothesis that a uniform least-cost flaw repair strategy could be highly effective in reducing the size of the search space in POCL planning. In those experiments, we compared LCFR against four other strategies: UCPOP, DUnf, and DUnf-LC, as defined above, and a new strategy, UCPOP-LC which we previously called LCOS (Joslin & Pollack, 1994):

**UCPOP-LC**      {n,s}LIFO / {o}LC

We included UCPOP-LC to help verify that search-space reduction results from a preference for flaws with minimal repair costs. If this is true, then UCPOP-LC ought to generate a smaller search space then DUnf, even though it does not delay any threats. Our results were as expected. UCPOP and DUnf, which do not do least-cost selection of open conditions, generated the largest search spaces; UCPOP-LC generated significantly smaller spaces; and DUnf-LC and LCFR generated the smallest spaces.

At the same time, we observed that LCFR incurred an unwieldy overhead, often taking longer to solve a problem than UCPOP, despite the fact that it was searching far fewer nodes. In part this was due a particularly inefficient implementation of LCFR that we were using, but in part it resulted from the fact that computing repair costs is bound to take more time than simply popping a stack (as in a LIFO strategy), or finding a flaw of a particular type (as in a strategy that prefers threats). We therefore explored approximation strategies, which reduce the overhead of flaw selection by accepting some inaccuracy in the repair cost calculation. For example, we developed the "Quick LCFR" (or QLCFR) strategy, which calculates the repair cost of any flaw only once, when that flaw is first encountered. In any successor node in which the flaw remains unresolved, QLCFR assumes that its repair cost has not changed. Our experiments with QLCFR showed it to be a promising means of making a least-cost approach sufficiently fast to pay for its own overhead. Additional approximation strategies were studied by Srinivasan and Howe (1995), who experimented with three variations of LCFR, along with a fourth, novel strategy that moves some of the control burden to the user.

---

6. When planning problems are cast as CSPs in the planner Descartes (Joslin & Pollack, 1996; Joslin, 1996), this correspondence is even more direct.

## 3.3 Threat Delays Revisited

Recently, Gerevini and Schubert (1996) have revived the idea that a flaw selection strategy should treat open conditions and threats differently, and have suggested that LIFO should be used as the tie-breaking strategy for deciding among open conditions. They combine these ideas in their ZLIFO strategy:

**ZLIFO** $\quad$ {n}LIFO / {o}$_0$LIFO / {o}$_1$New / {o}$_{2\text{-}\infty}$LIFO / {s}LIFO

The ZLIFO strategy gives highest priority to nonseparable threats, and then to forced open conditions. If there are neither nonseparable threats nor forced open conditions, ZLIFO will select an open condition using LIFO. It defers all separable threats to the end of the planning process. The name ZLIFO is intended to summarize the overall strategy. The "Z" stands for "zero-commitment", indicating that preference is given to forced open conditions: in repairing these, the planner is not making any commitment beyond what *must* be made if the node is ultimately to be refined into a complete plan. The "LIFO" indicates the strategy used for selecting among unforced open conditions.

For open conditions with a repair cost of exactly one, the ZLIFO strategy uses a tie-breaking strategy here called "New". It prefers the repair of an open condition that can only be established by introducing a new action over the repair of an open condition that can only be established by using an element of the start state. Gerevini and Schubert state that this preference "gave improvements in the context of Russell's tire changing domain ... without significant deterioration of performance in other domains" (1996, p. 104). However the difference was apparently not dramatic, and Gerevini believes this to be an implementation detail, though is open to the possibility that further study might show this preference to be significant (Gerevini, 1997).

Gerevini and Schubert make three primary claims about ZLIFO:

1. A POCL planner using ZLIFO will tend to generate a smaller search space than one using a pure LIFO strategy.

2. The reduction in search space using ZLIFO, relative to LIFO, is correlated with the complexity of the planning problem (where complexity is measured by the number of nodes generated by the pure LIFO strategy).

3. ZLIFO performs comparably with LCFR on relatively easy problems, and generates a smaller search space on harder problems.

The first two claims are consistent with what we found in the earlier LCFR studies. While a LIFO strategy pays no attention to repair costs, ZLIFO does, at least indirectly, both in its initial preference for nonseparable threats, which have a repair cost of no more than two, and in its secondary preference for forced opens.

The third claim is harder to square with our earlier LCFR study, in which the LIFO-based strategies, such as UCPOP and DUnf, generated much larger search spaces than the least-cost based strategies. What explains ZLIFO's performance? Gerevini and Schubert answer this question as follows:

> Based on experience with search processes in AI in general, [a LIFO] strategy has much to recommend it, as a simple default. In the first place, its overhead cost is low compared to strategies that use heuristic evaluation or lookahead to prioritize goals. As well, it will tend to maintain focus on the achievement of a particular higher level goal by regression ... rather than attempting to achieve multiple goals in a breadth-first fashion. [p. 103]

Their point about overhead is an important one. ZLIFO is a relatively inexpensive control strategy, and a competing strategy that does a better job of pruning the search space may end up paying excessive overhead. But it is the second point that addresses the question we are asking here, namely, how ZLIFO could produce *smaller* search spaces. Gerevini and Schubert go on to say that:

> [m]aintaining focus on a single goal should be advantageous at least when some of the goals to be achieved are independent. For instance, suppose that two goals G1 and G2 can both be achieved in various ways, but choosing a particular method of achieving G1 does not rule out any of the methods of achieving G2. Then if we maintain focus on G1 until it is solved, before attempting G2, the total cost of solving both goals will just be the sum of the costs of solving them independently. But if we switch back and forth, and solutions of both goals involve searches that encounter many dead ends, the combined cost can be much larger. This is because we will tend to search any unsolvable subtree of the G1 search tree repeatedly, in combination with various alternatives in the G2 search tree .... [p. 103]

This is certainly a plausible explanation. A key remaining question, of course, is the extent to which this explanation carries over to the many planning problems that involve interacting goals.

## 4. Experimental Comparison of Flaw Selection Strategies

As discussed in the previous section, several different proposals have been made in the literature about how best to reduce the size of the search space during POCL planning. These include:

- giving preference to threats over open conditions;

- giving preference only to certain kinds of threats (either separable or forced threats), and delaying other threats until after all open conditions have been resolved;

- giving preference to flaws that have minimal repair cost;

- giving preference to the most recently introduced flaws.

Moreover, different strategies have combined these preference schemes in different ways, and apparently conflicting claims have been made about the effects of these preferences on search-space size.

To resolve these conflicts, we performed experimental comparisons of POCL planners using a variety of flaw selection strategies. We gave particular attention to the comparison of LCFR and ZLIFO, because of the their apparently conflicting claims. LCFR generates its search space treating all flaws uniformly, using a least-cost approach to choose among them. ZLIFO distinguishes between flaw types (non-separable threats, open conditions, and separable threats), and uses a modified LIFO approach to select among the flaws in each class. The original LCFR studies would have led us to predict that ZLIFO would generate larger search spaces than did LCFR, but Gerevini and Schubert found just the opposite to be true. We aimed, then, to explain this discrepancy.

Our principal focus was on search-space size, for two reasons. First, the puzzle raised by LCFR and ZLIFO is one of space, not time. As we mentioned earlier, it is easy to see why ZLIFO would be faster than LCFR, even on a per node basis. A least-cost strategy must compute repair costs, while ZLIFO need only pop a stack containing the right type of flaws. The puzzle for us was not why ZLIFO was faster, but why it generated smaller search spaces. Second, we believe that understanding the effect of search control strategies on search-space size can lead to development of approximation techniques that produce speed-up as well; the QLCFR strategy (Joslin & Pollack, 1994) and Srinivasan and Howe's strategies (1995) are examples of this.

However, a secondary goal was to analyze the time requirements of the strategies we compared, and we therefore collected timing data for all our experiments. As we discuss in Section 4.6, the strategy that tends to generate the smallest search space achieves enough of a reduction to pay for its own overhead, by and large.

## 4.1 Experimental Design

To conduct our comparison, we implemented a set of flaw selection strategies in UCPOP v.4.[7] Table 1 lists the strategies that we implemented. Except for LCFR-DSep and DUnf-Gen, which are discussed later, all the implemented strategies were described in Section 3.

We tested all the strategies on three problem sets, also used in our earlier work (Joslin & Pollack, 1994) and in Gerevini and Schubert's (1996):

1. *The Basic Problems*, 33 problems taken from the test suite distributed with the UCPOP system. These include problems from a variety of domains, including the blocks world, the Monkeys and Bananas problem, Pednault's (1988) briefcase-and-office problem, Russell's (1992) tire changing world, etc.

---

7. Note that the experiments in both our earlier LCFR paper (Joslin & Pollack, 1994) and Gerevini and Schubert's (1996) ZLIFO paper were run using an earlier version (v.2) of UCPOP. As a result, the number of nodes produced in our experiments sometimes differs from what is reported in these other two papers. This appears to be largely due to the fact that UCPOP v.4 puts the elements of a new set of open conditions onto the flaw list in the reverse order of the way in which UCPOP v.2 does (Gerevini, 1997). As discussed below in Sections 4.3–4.5, we studied the influence of this ordering change by also collecting data using a modified version of UCPOP v.4 in which we reversed the order of conditions entered in the open list. While the resulting numbers are similar to those previously published, they are not identical, leading us to conclude that there are additional subtle differences between v.2 and v.4. However, because *all* the experiments on which we report here were run using the same version of UCPOP, we believe this to be a fair comparison of the strategies.

| UCPOP | {n,s}LIFO / {o}LIFO |
|---|---|
| UCPOP-LC | {n,s}LIFO / {o}LC |
| DSep | {n}LIFO / {o}LIFO / {s}LIFO |
| DSep-LC | {n}LIFO / {o}LC / {s}LIFO |
| DUnf | {n,s}$_0$LIFO / {n,s}$_1$LIFO / {o}LIFO / {n,s}$_{2-\infty}$LIFO |
| DUnf-LC | {n,s}$_0$LIFO / {n,s}$_1$LIFO / {o}LC / {n,s}$_{2-\infty}$LIFO |
| DUnf-Gen | {n,s,o}$_0$LIFO / {n,s,o}$_1$LIFO / {n,s,o}$_{2-\infty}$LIFO |
| LCFR | {o,n,s}LC |
| LCFR-DSep | {n,o}LC / {s}LC |
| ZLIFO | {n}LIFO / {o}$_0$LIFO / {o}$_1$New / {o}$_{2-\infty}$LIFO / {s}LIFO |

Table 1: Implemented Flaw Selection Strategies

2. *The Trains Problems*, three problems taken from the TRAINS transportation domain (Allen, Schubert, & *et al.*, 1995).

3. *The Tileworld Problems*, seven problems taken from the Tileworld domain (Pollack & Ringuette, 1990).

We ran each strategy on each problem twice. The first time, we imposed a node limit, of 10,000 nodes for the basic problems, and of 100,000 nodes for the Trains and Tileworld problems. The second time, we imposed a time limit, of 100 seconds for the basic problems, and of 1000 seconds for the Trains and Tileworld problems.

Gerevini and Schubert experimented with several different node selection strategies for the Trains and Tileworld domains, so to facilitate comparison we also used the same node selection strategies as they did. For the basic problems, we used $S + OC$.

In reporting our results, we make use not only of raw counts of nodes generated and computation time in seconds taken, but we also compute a measure of how badly a strategy performed on a given problem or set of problems. We call this measure %-overrun, and compute it as follows. Let $m$ be the minimum node count on a given problem for *any* of the strategies we tested, and let $c$ be the node count for a particular strategy $S$. Then

$$\text{\%-overrun}(S) = [(c - m)/m] * 100$$

Thus, for example, if the best strategy on a given problem generated 100 nodes, then a strategy that generated 200 nodes would have a 100 %-overrun on that problem. The strategy that does best on a given problem will have a %-overrun of 0 on that problem. In Section 4.6, we make use of similarly computed %-overruns for computation time.

If a strategy hit the node limit, we set $c$ to the relevant node limit (10,000 or 100,000) to compute its node-count %-overrun.[8] Similarly, if a strategy hit the time limit, we used the relevant time limit (100 or 1000) to compute the computation-time %-overrun.

---

8. Because of the way in which UCPOP completes its basic iteration, it sometimes will go somewhat beyond the specified node limit before terminating the run. In such cases, we used the node limit value, rather than the actual number of nodes generated, in our computation of %-overrun.

Online Appendix A provides the raw data—node counts and computation-time taken—for all the experiments we conducted; it also includes computed %-overruns.

In conducting experiments such as these, one has to set either a node- or time limit cutoff for each strategy/problem pair. However, there is always a danger that these cutoffs unfairly bias the data, if the limits are set in such a way that certain strategies that fail would instead have succeeded were the limits increased slightly. We have carefully analyzed our data to help eliminate the possibility of such a bias; details are given in Appendix A.

## 4.2 The Value of Least-Cost Selection

Having described our overall experimental design, we now turn to the analysis of the results. To begin, we sought to re-establish the claims we originally made in our earlier work. Specifically, we wanted first to reconfirm, using a larger data set, that least-cost flaw selection is an effective technique for reducing the size of the search space generated by a POCL planner. We therefore ran an experiment in which we compared the the node counts for the five strategies we had earlier studied—LCFR, DUnf, DUnf-LC, UCPOP, and UCPOP-LC—plus one new one, DUnf-Gen, explained below.

The results of this experiment are shown in Figures 2 and 3. The former is a log-log scatter plot, showing the performance of each of the six strategies on the 33 problems in the basic set. The problems were sorted by the minimal number of nodes generated on them by any of the six strategies. Thus, the left-hand side of the graph includes the problems that at least one of the six strategies found to be relatively easy, while the right-hand side has the problems that were hard for all six strategies. We omitted problems that none of the six strategies were able to solve. The actual number of nodes generated by each strategy is plotted on the Y-axis, against the minimal number of nodes for that problem, on the X-axis. LCFR's performance is highlighted with a line connecting its data points. This graph shows that, in general, LCFR generates small search spaces on this problem set, relative to the other strategies in this class. There were only six problems for which LCFR was not within 10% of the minimum. Three of these are in the Get-Paid/Uget-Paid class of problems—including two of the "hardest" problems (UGet-Paid3 and UGet-Paid4). We discuss this class of problems more in Section 4.5.

An alternative view of the data is given in Figure 3, which shows the aggregate performance of the six strategies, i.e., the average of their node-count %-overrun on the basic problems. As can be seen, LCFR has the smallest average %-overrun.

Figures 4 and 5 present similar views of the data for the Tileworld domain, while Figure 6 gives the data for the Trains problems. On the Trains domain, these six strategies were only able to solve the easiest problem (Trains1), so we simply show the actual node counts in Figure 6. We have omitted two data points, because they were so extreme that their inclusion on the graph made it impossible to see the differences among the other strategies: LCFR and DUnf-Gen with $S + OC + UC$ node selection took 28,218, and 35,483 nodes, respectively, to solve the problem.

For the Tileworld and Trains problems, we observed the same sorts of interactions between node and flaw selection strategies as were seen by Gerevini and Schubert. Specifically, LCFR performs relatively poorly with $S + OC$ on the Tileworld problems, and it performs very poorly with $S + OC + UC$ on the Trains problems. However, when paired with the other node-selection strategies, LCFR produces the smallest search spaces of any strategies in this class.

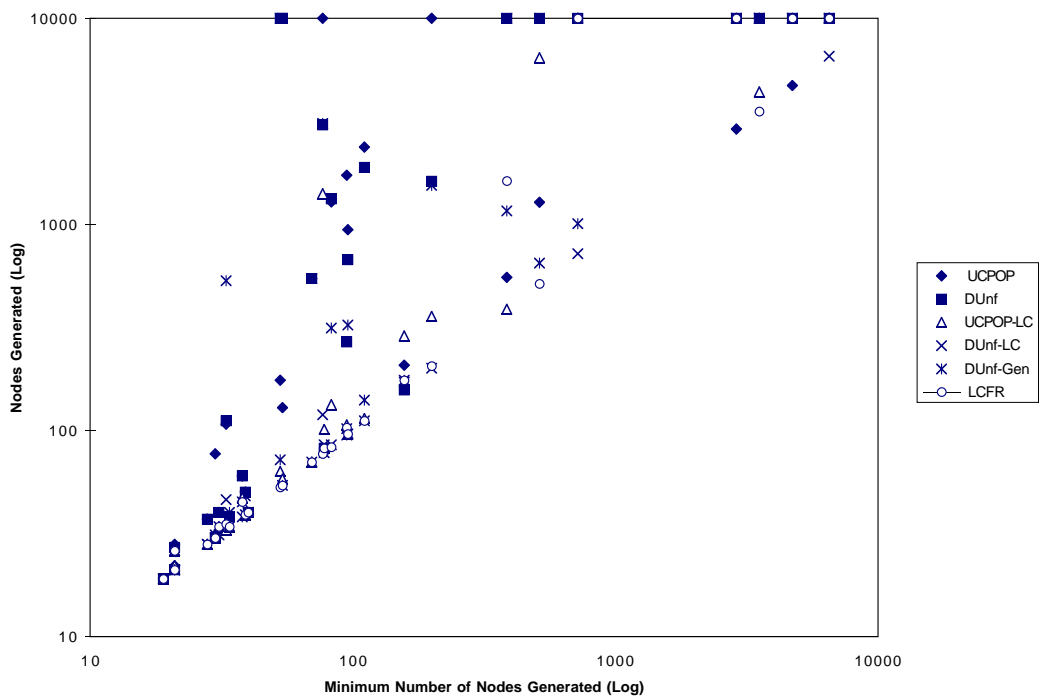

Figure 2: Basic Problems: Node Counts for Strategies without Forced-Flaw Delay

In sum, LCFR does tend to produce smaller search spaces than the other strategies in this class. But a question remains. LCFR uses a least-cost strategy, and a side effect of this is that it will prefer forced flaws, since forced flaws are low-cost flaws. It is therefore conceivable that LCFR's performance is mostly or even fully due to its preference for forced flaws, and not (or not greatly) influenced by its use of a least-cost strategy for unforced flaws. This same hypothesis could explain why DUnf-LC consistently outperforms DUnf, and why UCPOP-LC consistently outperforms UCPOP.

It was to address this issue that we included DUnf-Gen in our experiment. DUnf-Gen is a simple strategy that prefers forced flaws of any kind, and otherwise uses a LIFO regime. We would expect DUnf-Gen and LCFR to perform similarly, since they frequently make the same decision. Specifically, they will both select the same flaw in any node in which there is a forced flaw; they will differ when there are only unforced flaws, with DUnf-Gen selecting a most recently introduced flaw and LCFR selecting a least-cost flaw.

In practice, DUnf-Gen's performance closely mimicked that of LCFR's. On the basic problem set it did only marginally worse than LCFR. In fact, it does marginally *better* when we reverse the order in which the planner adds the preconditions of each new step to the open list (see Section 4.4). LCFR does somewhat better than DUnf-Gen on both the Trains and Tileworld problems, and this is true regardless of the order in which the preconditions were added to the open list, but the extent to which it does better varies.

Thus, the data is inconclusive about the value of using a least-cost strategy for unforced flaws. LCFR clearly benefits from selecting forced flaws early (as a side effect of preferring least-cost flaws), but it *may* not matter whether it continues to use a least-cost strategy for the unforced flaws. If indeed it is generally sufficient to use a least-cost strategy only for forced flaws, then ZLIFO's performance is somewhat less puzzling, since ZLIFO also prefers forced flaws. However the puzzle is not completely resolved. After all, DUnf-Gen, like ZLIFO, prefers forced flaws and and then makes LIFO-based decisions about unforced flaws, and while its performance is not clearly inferior to LCFR's, neither is it clearly superior. Even if the use of LIFO for unforced flaws does not obviously increase the search-space, neither does it appear to decrease it.

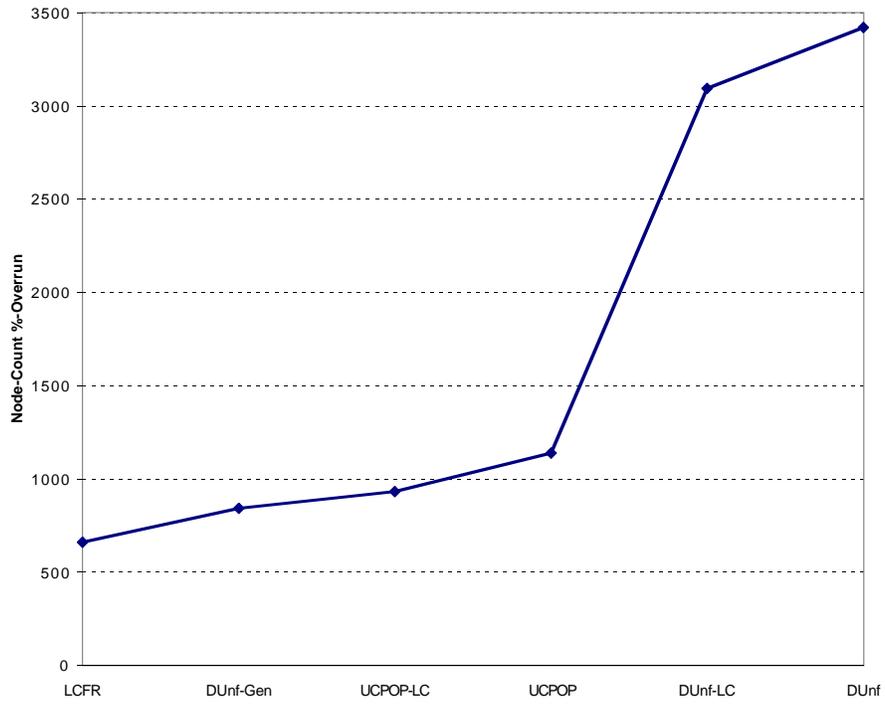

Figure 3: Basic Problems: Aggregate Performance for Strategies without Forced-Flaw Delay

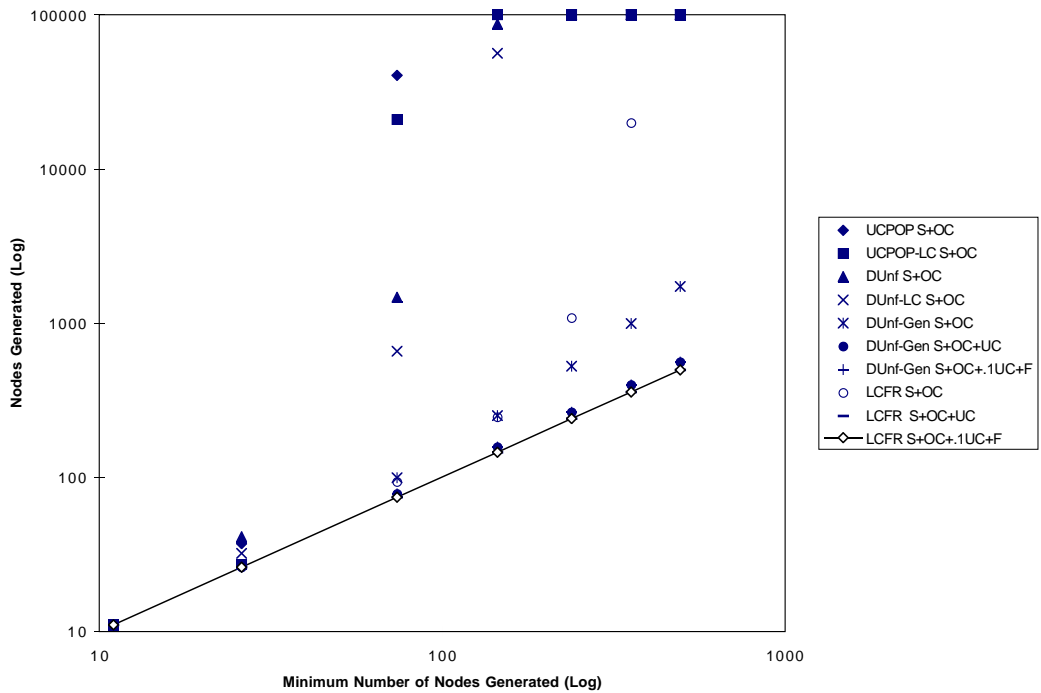

Figure 4: Tileworld Problems: Node Counts for Strategies without Forced-Flaw Delay

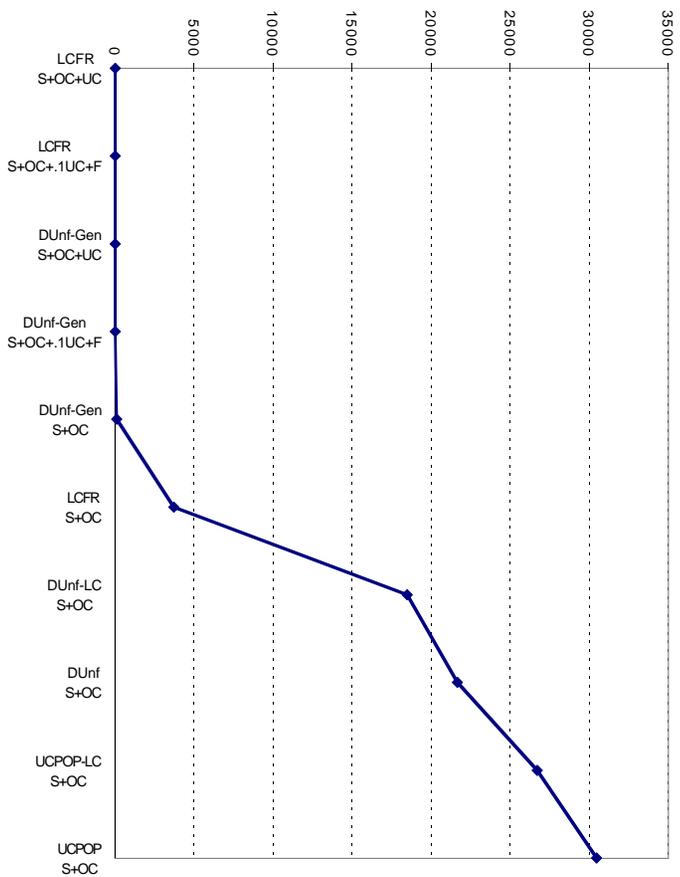

Figure 5: Tileworld Problems: Aggregate Performance for Strategies without Forced-Flaw Delay

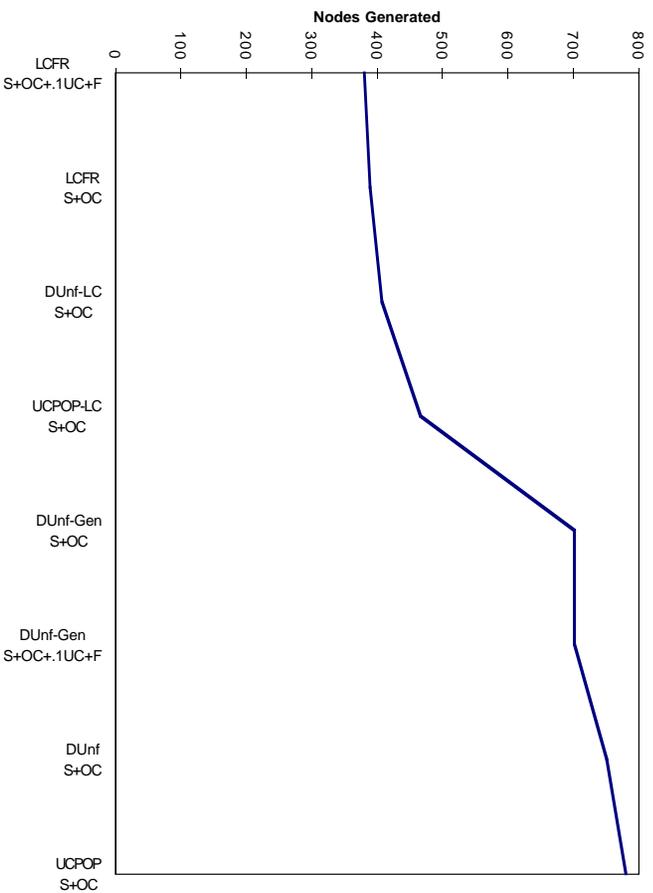

Figure 6: Trains 1: Node Counts for Strategies without Forced-Flaw Delay

## 4.3 Comparing LCFR and ZLIFO

We next turn to a direct comparison of LCFR and ZLIFO. Gerevini and Schubert compared these strategies on only a few problems. To get a more complete picture of the performance of both LCFR and ZLIFO, we ran them both on all the problems from our three problem sets.

The data for the basic problem set is shown in Figure 7. We have sorted the problems by the difference between the node counts produced by LCFR and ZLIFO. Thus, problems near the left-hand side of the graph are those for which LCFR generated a smaller search space, while problems near the right-hand side are the ones on which ZLIFO had a space advantage. We omit problems which neither strategy could solve.

As can be seen, on some problems (notably R-Test2, Move-Boxes, and Monkey-Test2), LCFR generates a much smaller search space than ZLIFO, while on other problems (notably Get-Paid4, Hanoi, Uget-Paid4, and Uget-Paid3), ZLIFO generates a much smaller search space. These are problems on which LCFR also did worse than the strategies mentioned above in Section 4.2.

As we noted earlier, one of the major changes between UCPOP v.2 and v.4 is that v.4 puts the elements of a new set of open conditions onto the flaw list in the reverse order from that of v.2. This ordering may make a difference, particularly for LIFO-based strategies. Indeed, other researchers have suggested that one reason a LIFO-based strategy may perform well is because it can exploit the decisions made by the system designers in writing the domain operators, since it is in some sense natural to list the most constraining preconditions of an operator first (Williamson & Hanks, 1996). We therefore also collected data for a modified version of UCPOP, in which the preconditions for each step are entered onto the open condition in the reverse of the order in which they would normally be entered. We discuss the results of this modification in more detail in the next two sections, but for now, we simply present the node counts for LCFR and ZLIFO with the reversed precondition insertion, in Figure 8. As can be seen, there are a few problems on which reversing the precondition ordering has a significant effect (notably FIXB and MonkeyTest2), but by and large LCFR and ZLIFO showed the same relative performance.

For the problems in the basic set, it is difficult to discern an obvious pattern of performance. In contrast to what Gerevini and Schubert suggest, there does not seem to be a clear correlation between the difficulty of the problem, measured in terms of nodes generated, and the relative performance of LCFR and ZLIFO. (In fact, it is a little difficult to determine which strategy's node-count should serve as the measure of difficulty.) On the other hand, it is true that in the aggregate, ZLIFO generates smaller search spaces than LCFR on the basic problems. With the default precondition ordering, ZLIFO obtains an average %-overrun of 212.62, while LCFR obtains 647.57. With reverse ordering, ZLIFO's average %-overrun is 244.24, while LCFR's is 914.87. The fact that LCFR's relative performance is worse when the preconditions are entered in the reverse direction results primarily from its failure on MonkeyTest2 in the reverse direction.

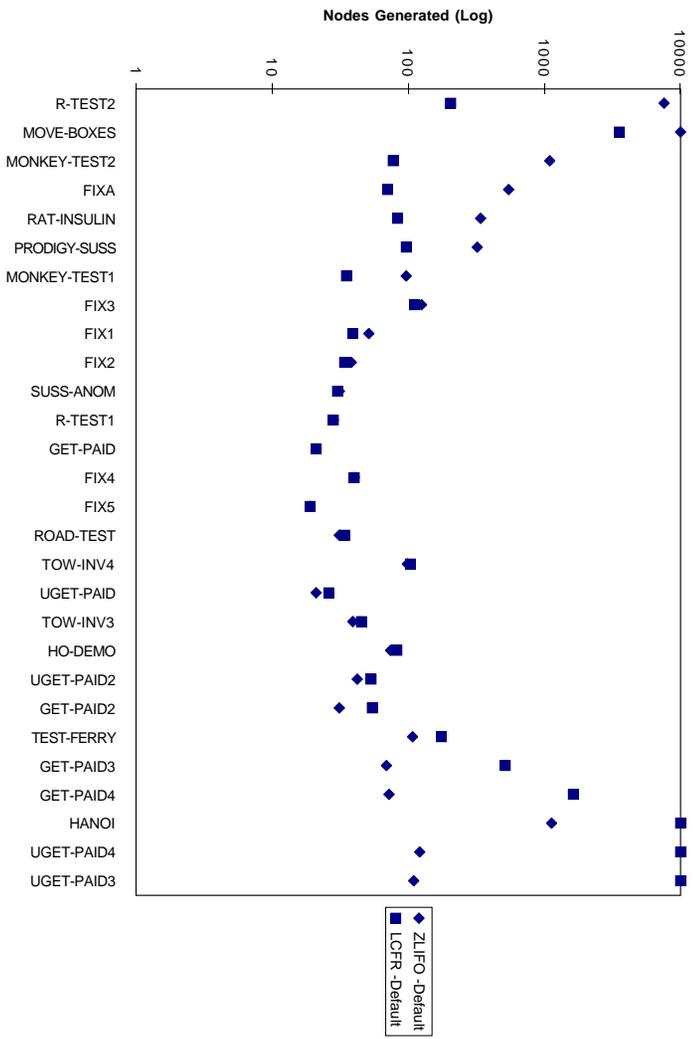

Figure 7: Basic Problems: Node Counts for LCFR and ZLIFO

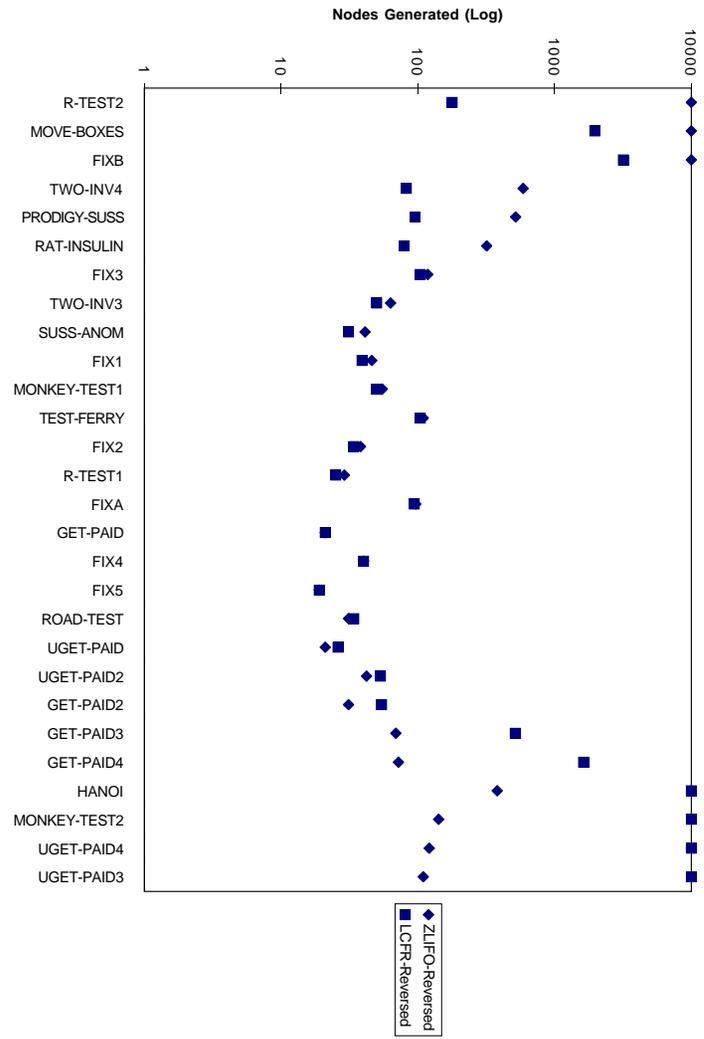

Figure 8: Basic Problems: Node Counts for LCFR and ZLIFO with Reversed Precondition Insertion

The Trains data is scant. Neither LCFR nor ZLIFO can solve the hardest problem, Trains3, regardless of whether the preconditions are entered in the default or the reverse order. (In fact, none of the strategies we studied were able to solve Trains3.) But, at least when the preconditions are entered in the default order, ZLIFO can solve Trains2, and LCFR cannot. With reverse precondition insertion, neither strategy can solve Trains2. The data are shown in Figure 9. Note that LCFR's performance is essentially the same for both node-selection strategies shown.

Finally, the Tileworld data, for the default order of precondition insertion, is shown in Figure 10. Here is the only place in which LCFR clearly generates smaller search spaces than ZLIFO. We have not also plotted the data for reverse precondition insertion, because most of the strategies are not affected by this change. There is however, one very notable exception: with reversed insertion, ZLIFO (with $S + OC + .1UC + F$) does much better—indeed, it does as well as LCFR. We return to the influence of precondition ordering on the Tileworld problems in Section 4.5.

For now, however, it is enough to observe that our experiments show that ZLIFO does tend to generate smaller search spaces than LCFR. It does so on the basic problem set, regardless of the order of precondition insertion, it does so on Trains for one ordering (and does no worse than LCFR on the other ordering), and it does as well as LCFR for the Tileworld problems when the preconditions are inserted in the reverse order. The only exception is the Tileworld problem set when the preconditions are inserted in default order: there LCFR does better.

## 4.4 The Value of Separable-Threat Delay

Our first two analyses were essentially aimed at replicating earlier results from the literature, namely the LCFR results and the ZLIFO results. We next address the question of how to square these results with one another.

Recall that LCFR and ZLIFO differ in two key respects. First, LCFR treats all flaws uniformly, while ZLIFO distinguishes among flaw types, giving highest preference to non-separable threats, medium preference to open conditions, and lowest preference to separable threats. Second, while LCFR uniformly makes least-cost selections, ZLIFO uses a LIFO strategy secondary to its flaw-type preferences (but after giving preference to forced open conditions). The comparisons made in Section 4.2 suggest that the use of a LIFO strategy for unforced flaws should at best make little difference in search-space size, and may possibly lead to to the generation of larger search spaces. On the other hand, the first difference presents an obvious place to look for a relative advantage for ZLIFO. After all, what ZLIFO is doing is delaying separable threats, and Peot and Smith demonstrated the effectiveness of that approach in their DSep strategy.

Peot and Smith's proof that DSep will never generate a larger search space than UCPOP does not transfer to LCFR. There are planning problems for which LCFR will generate a smaller search space than DSep. Their proof relies on the fact that, in DSep, open conditions will be selected in the same order, regardless of when threats are selected. But the selection of a threat in LCFR can influence the repair cost of an open condition (e.g., by promoting an action so that it is no longer available as a potential establisher for some condition), and this in turn can affect the order in which the remaining open conditions are selected.

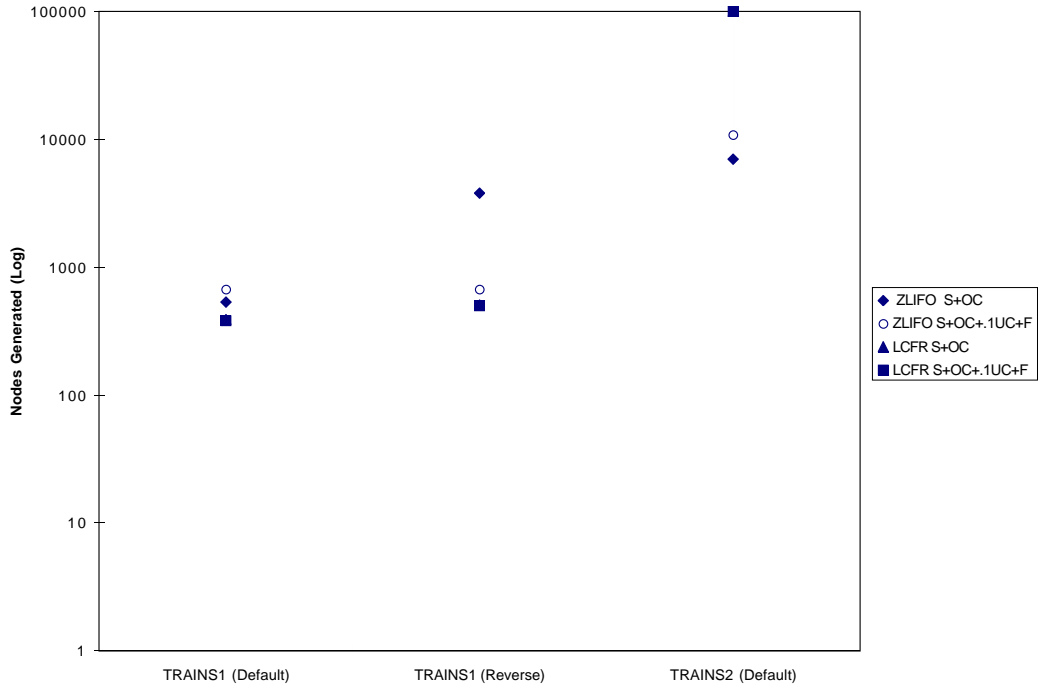

Figure 9: Trains Problems: Node Counts for LCFR and ZLIFO

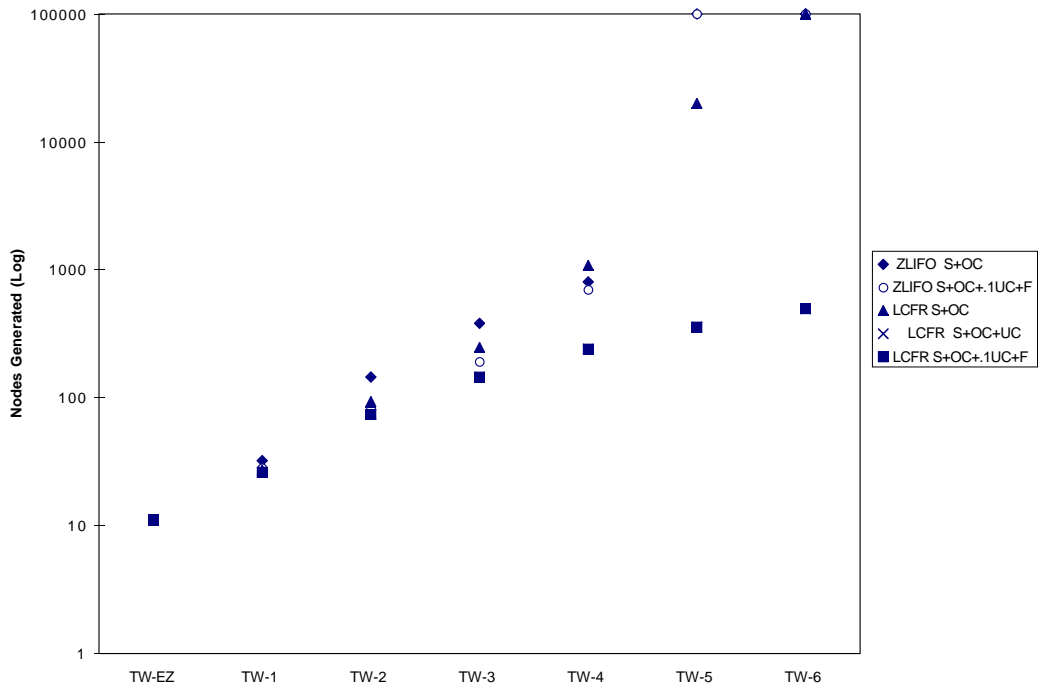

Figure 10: Tileworld Problems: Node Counts for LCFR and ZLIFO

Nonetheless, despite the fact that one can't *guarantee* that delaying separable threats will lead to a reduction in search-space size, the motivation behind DSep is still appealing: separable threats may often simply disappear during subsequent planning, which will naturally lead to a reduction in search-space size. For this reason, we implemented a slightly modified version of LCFR, which we called LCFR-DSep, in which separable threats are delayed. Note that it is relatively easy to do this in the UCPOP system, which provides a switch, the *dsep* switch, which when turned on will automatically delay the repair of all separable threats. As defined earlier in Table 1, the definition of LCFR-DSep is:

**LCFR-DSep**     {n,o}LC / {s}LC

Our hypothesis was that if ZLIFO's reduction in search-space size were largely due to its incorporating a DSep approach, then LCFR-DSep ought to be "the best of both worlds", combining the advantages of LCFR's least-cost approach with the advantages of a DSep approach.

On the basic problems, LCFR-DSep proved to have the smallest average node-count %-overrun of on the basic problems of *all* of the strategies tested. Moreover, this was true even when we reversed the order in which the preconditions of an operator were added to the open list. Figure 11 gives the average node-count %-overruns for both the unmodified UCPOP v.4 (labeled "default") and the modified version in which we reversed the precondition ordering (labeled "reverse"). Reversing the ordering does not effect the conclusion that LCFR-DSep generates the smallest search spaces for these problems; in fact, in general it had very little affect on the relative performance of the strategies at all. The only notable exception, which we mentioned earlier, is that the relative performance of LCFR and DUnf-Gen flips.

For more detailed comparison, we plot node counts on the basic problems for LCFR, ZLIFO, and the Separable-Threat Delay strategies in Figure 12. For ease of comparison, we again show the data sorted by the difference between LCFR and ZLIFO's node counts. The problems near the left-hand side of the graph are, again, those for which LCFR generated a smaller search space than ZLIFO; the problems near the right are those for which it generated a larger search space. As can be seen, LCFR-DSep nearly always does as well as, or better than LCFR. It does much better than ZLIFO on the problems that LCFR is good at. And it also does *much* better than LCFR on the problems that ZLIFO is good at. However, ZLIFO still outperforms LCFR-DSep on this latter class of problems.

Another view of the data is given in Figure 13, the log-log scatter plot for the basic problems, for all the strategies we studied. This time we have highlighted LCFR-DSep's performance. Although there are some problems for which it does not produce a minimal search space, its performance on the individual problems is actually quite good, consistent with its good aggregate performance.

At least for the basic problems, augmenting the simple LCFR strategy with a delay of separable threats reduces the search space as expected. This in turn suggests that when LCFR generates a larger search space than ZLIFO, that is due in large part to the fact that

POLLACK, JOSLIN, & PAOLUCCI

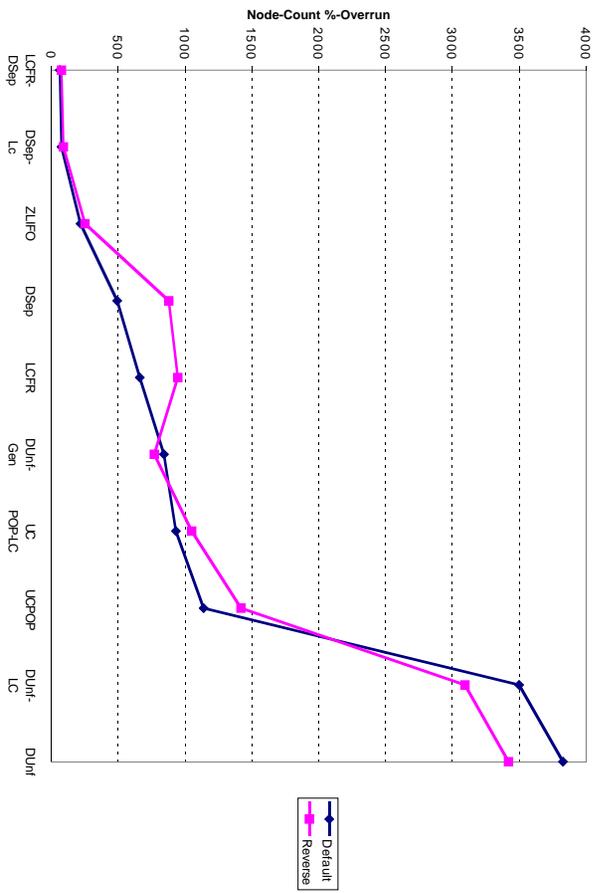

Figure 11: Basic Problems: Aggregate Performance for all Strategies

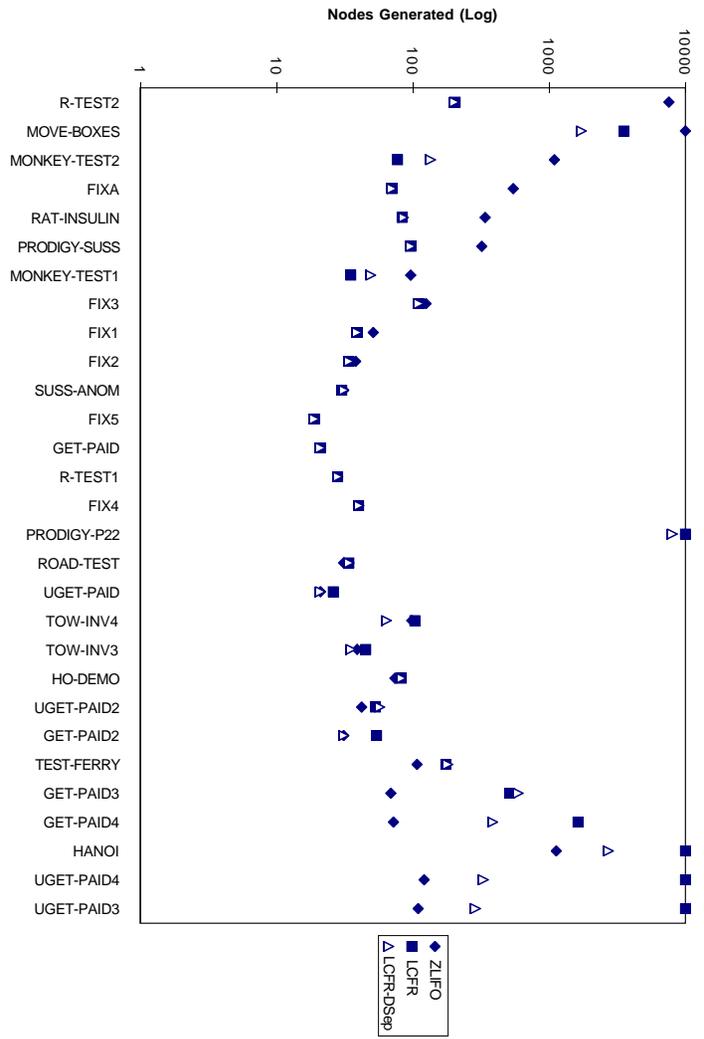

Figure 12: Basic Problems: Node Counts for LCFR, ZLIFO, and DSep Strategies

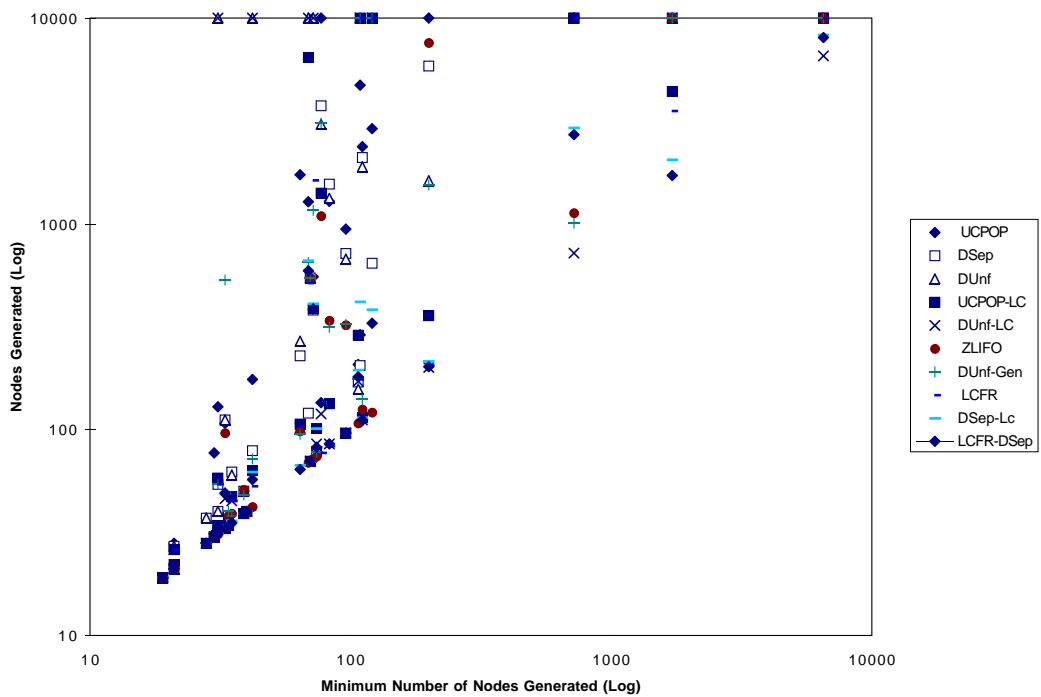

Figure 13: Basic Problems: Node Counts for all Strategies

it does not delay separable threats. ZLIFO's primary advantage relative to LCFR seems not to be its use of a LIFO strategy for unforced threats, but rather its separable-threat delay component. Combining separable-threat delay with a least-cost approach yields a strategy that tends to generate smaller search spaces than either strategy by itself for the basic problem set. However, analysis of the Trains and Tileworld problem sets reveals the situation to be a little more complicated than the comparison of the basic problems would suggest, as we discuss in the next section.

## 4.5 The Need for Domain Information

The Tileworld and Trains domains problems challenge overly simple conclusions we might draw from the basic problem sets. We consider each set of problems in turn.

### 4.5.1 The Tileworld Problems

The Tileworld domain involves a grid with tiles and holes, and the goal is to fill each hole with a tile. This goal can be achieved with a **fill** operator, which has two preconditions: the agent must be at the hole, and it must be holding a tile. In our encoding, an agent can hold up to four tiles at a time. The **go** operator is used to achieve the (sub)goal of being at a hole, while the **pickup** operator is used to achieve the (sub)goal of holding a tile. In the normal way, **go** has a precondition of being at some location, namely whatever location the agent will move from. **Pickup** has a precondition of being at the location of some tile. The problems in the Tileworld problem set differ from one another in the number of holes that the agent must fill: each problem adds another hole.

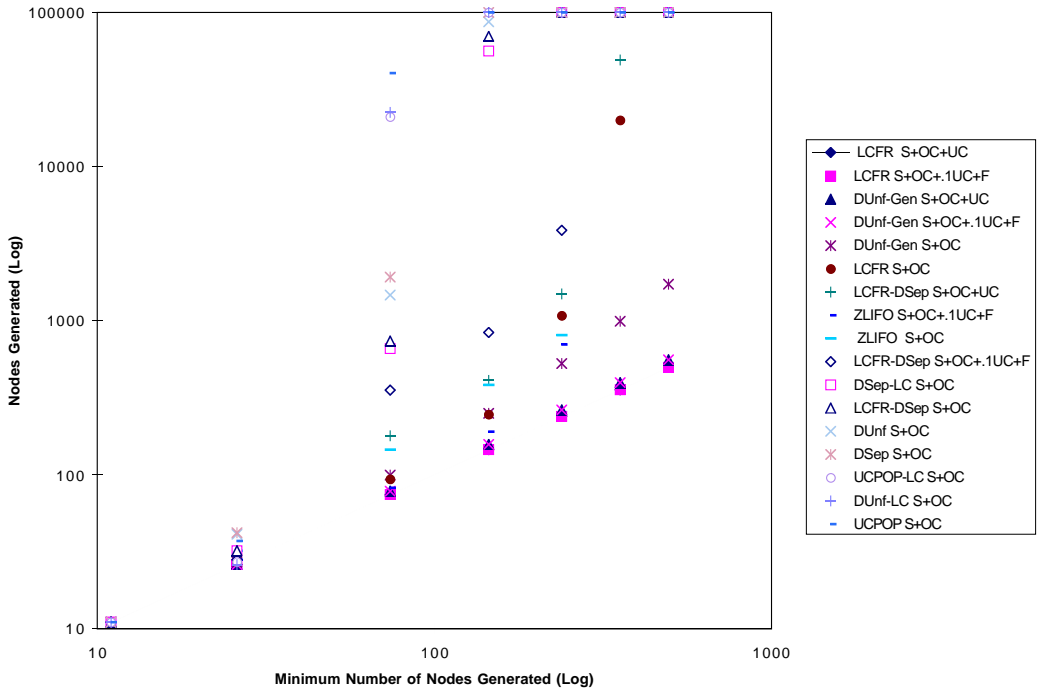

Figure 14: Tileworld Problems: Node Counts for All Strategies

Figures 14 gives the log-log plot for the various strategies on the Tileworld problems, when the preconditions were entered in the default order. Note that LCFR $(S + OC + UC)$ is the strategy highlighted. Three other strategies were almost indistinguishable from LCFR $(S + OC + UC)$, namely, LCFR $(S + OC + .1UC + F)$, DUnf-Gen $(S + OC + UC)$ and DUnf-Gen$(S + OC + .1UC + F)$. All the other strategies performed worse. This can more easily be seen in Figure 15, which gives the aggregate performance for the leading strategies: those that were able to solve all seven Tileworld problems. In fact, these leading strategies were able to solve the seven Tileworld problems without generating more than 1800 nodes for any problem. In contrast, the remaining strategies failed on at least one, and up to four, of the seven problems, given the limit of 100,000 nodes generated.

What was originally surprising to us is that on the Tileworld problems, delaying separable threats actually seems to hurt performance. The strategies that did best were those like LCFR and DUnf-Gen that do *not* delay separable threats. LCFR-DSep, ZLIFO, DSep-LC, and DSep all generated larger search spaces, in contrast to what we would have predicted given the experiments on the basic problem set.

To understand this result, we looked in detail at the planning trace for these problems. What that revealed is that for the Tileworld domain, the early resolution of separable threats has an important advantage: it imposes what turns out to be the correct temporal ordering between the steps of going to up a tile (to pick it up), and carrying it to a hole. Virtually all the strategies create subplans like the one shown in Figure 16. The goals involve filling holes, so the planners insert steps to go to and pick up a tile, and to go to the hole. At this point, there are two separable threats: (1) the effect of going to the hole, $\neg at(X)$, threatens the link between going to the tile and picking it up ($at(Z)$), and (2) the effect of going to the tile, $\neg at(W)$, threatens the link between going to the hole and filling it ($at(Y)$). Both threats are separable, because $X$ and $W$ will be unbound; the planner does not yet know where it will be traveling from. But there is only one valid temporal ordering that will resolve these threats: going to the tile must precede picking up the tile, which in turn must precede going to the hole. Once this temporal ordering is determined, further planning goes smoothly.

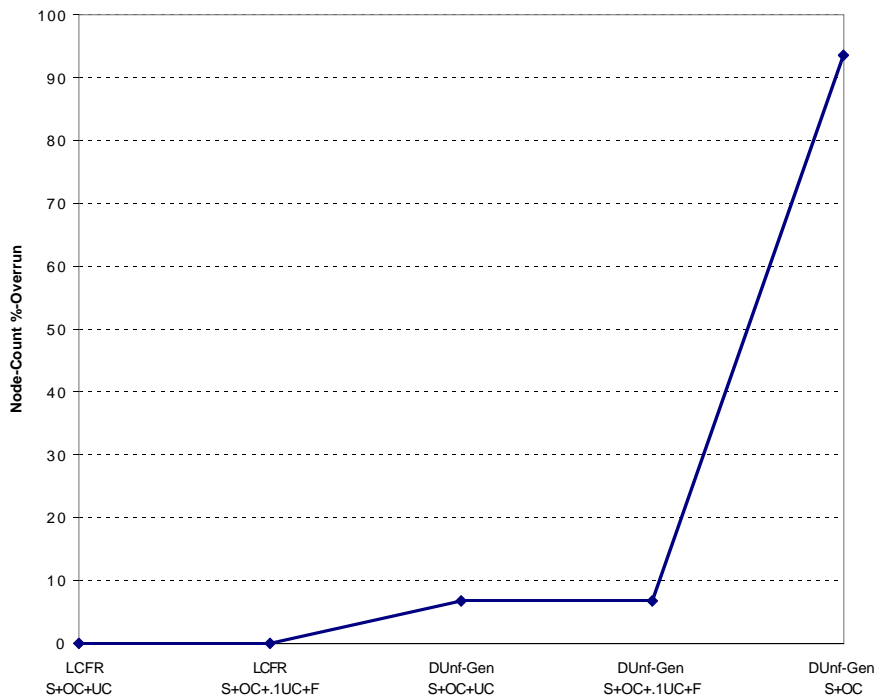

Figure 15: Tileworld Problems: Aggregate Performance for Leading Strategies

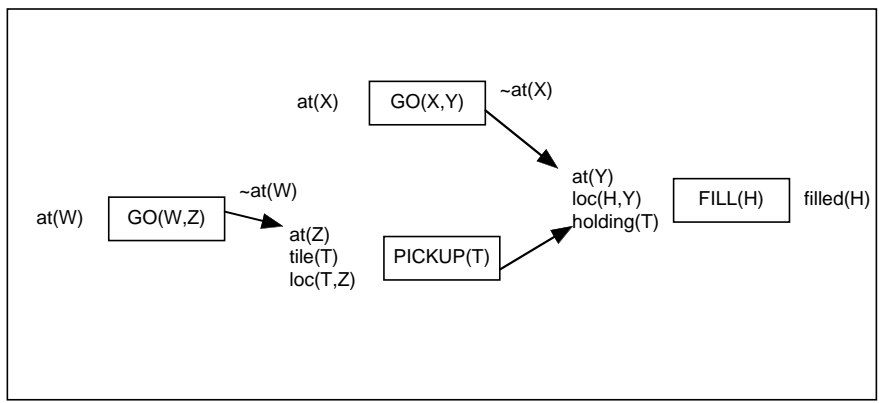

Figure 16: Typical Partial Plan for the Tileworld Domain

In contrast, if this ordering decision is not made, the planner can often "get lost", attempting to find plans in which it goes from some location to the hole and then from the hole to the tile. There are many ways to attempt this, because there are many different tiles to select, and many different locations to move among. The planner may try many of these alternatives before determining that there is a fundamental inconsistency in these plans, and that they are destined to fail. The larger the number of holes to be filled, the worse the situation becomes.

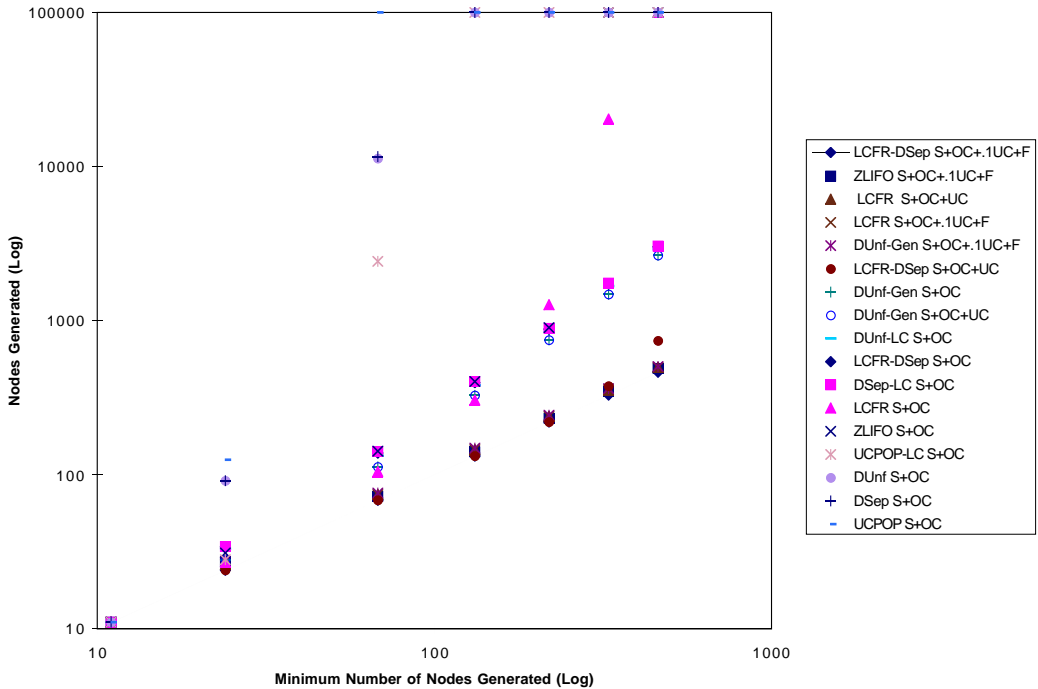

Figure 17: Tileworld Problems: Node Counts with Reversed Precondition Insertion

Sometimes the planner may make the right decision about temporal ordering even if it has deferred separable threats. When faced with the partial plan in Figure 16, if the planner does not select a threat, it will select from among several open conditions. It can attempt to establish the precondition of going to the hole $(at(X))$ by reusing the effect of going to the tile $(at(Z))$, or it can do the reverse, and attempt to establish the precondition of going to the tile $(at(W))$ by reusing the effect of going to the hole $(at(X))$. Of course, the first solution is the right one, and includes the critical temporal ordering constraint, while the second will eventually fail.

The order in which the open conditions are selected will determine which of these two choices the planner makes. When preconditions are entered in the default order, planners that delay separable threats end up making the latter, problematic choice. In contrast, when the preconditions are entered in the reverse order, the planners make what turns out to be the correct choice. Thus, for the experiments in which we reversed precondition insertion, we see a different pattern of performance, as shown in Figures 17–18.[9]

When the preconditions are entered in the reverse order, a larger number of strategies perform well, solving all the problems. In particular, with $S + OC + .1UC + F$ node-selection, the performance of LCFR, DUnf-Gen, ZLIFO, and LCFR-DSep is virtually indistinguishable. It is important to note that the leading strategies that do not delay separable threats—LCFR and DUnf-Gen—are not affected much by the reversal of precondition insertion for the Tileworld problems; in fact, LCFR's performance is identical in both cases. In contrast, the strategies that use separable-threat delay—LCFR-DSep, ZLIFO, and DSep-LC—all perform much better when we reverse precondition insertion. This is explained by our analysis above.

---

9. To preserve readability, in Figure 18, we have used "(1)" to denote $S + OC$, "(2)" for $S + OC + UC$, and "(3)" for $S + OC + UC + .1F$.

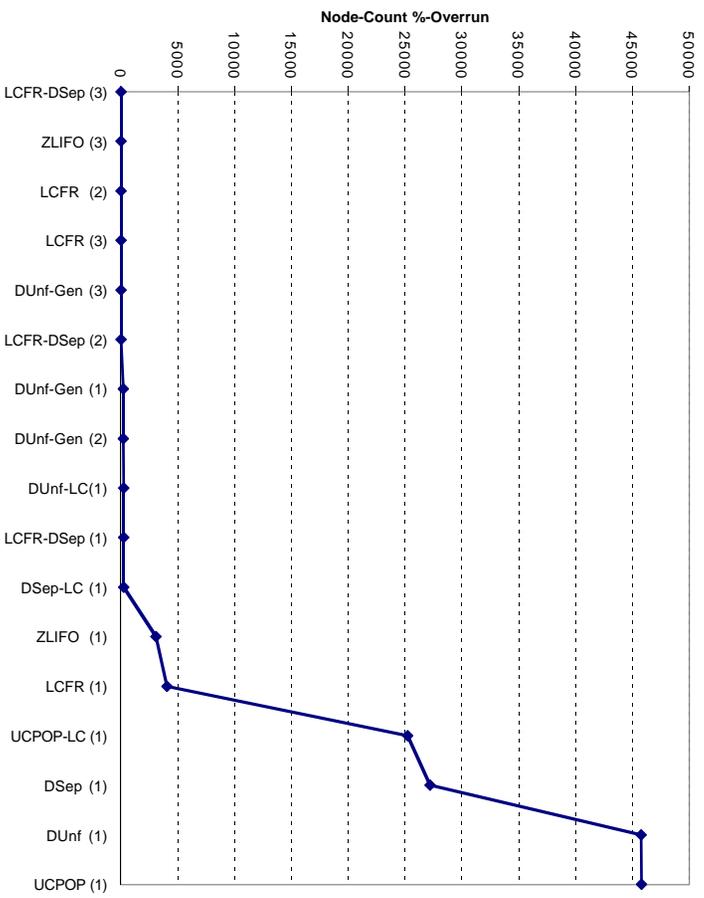

Figure 18: Tileworld Problems: Aggregate Performance for all Strategies with Reversed Precondition Insertion

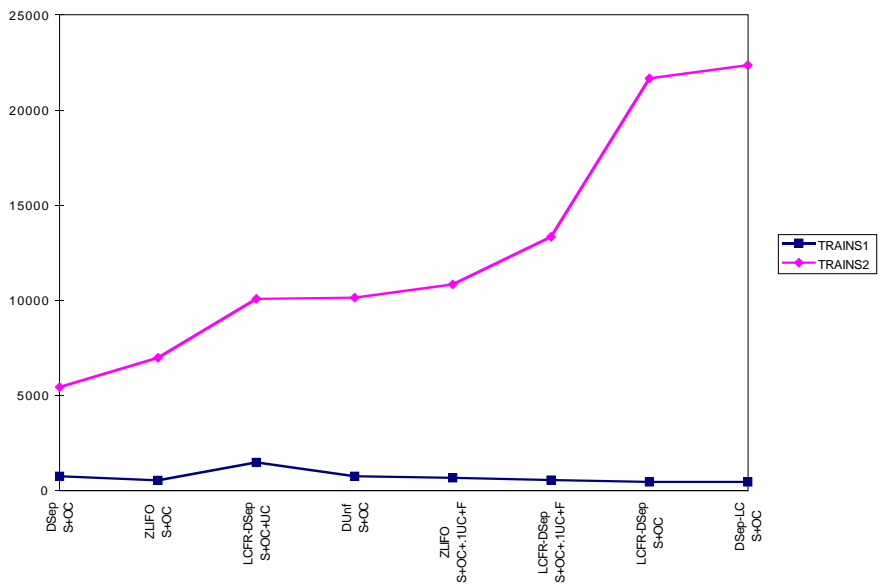

Figure 19: Trains Problems: Node Counts

In sum, what is most important for the Tileworld domain is for the planner to recognize, as early as possible, that there are certain required temporal orderings between some of the steps in any successful plan. Every successful plan will involve going to a tile before going to a hole, although there is flexibility in the order in which multiple holes are visited, and in the interleaving of picking up tiles and dropping them in holes. For the strategies we studied, there were two different methods that led to this temporal constraint being added to the plan. It was added when the planner selected a separable threat to resolve, and it was added when it selected one particular precondition to resolve before another.

### 4.5.2 THE TRAINS AND GET-PAID PROBLEMS

The Trains domain present a somewhat different variation on our original conclusions. The Trains domain involves a set of locations and objects, and the goal is to transport various objects from specific starting locations to specified destinations. Gerevini and Schubert studied three Trains problems. All of our strategies failed to successfully complete the hardest of these (Trains3) within either the 100,000 node or the 1000 second limit. Moreover, many of them also failed on the second hardest (Trains2). Caution must therefore be taken in interpreting the results, as there are a limited number of data points.

Figure 19 gives the node counts for the Trains domain, with preconditions inserted in the default order. We only show the strategies that were able to solve both Trains1 and Trains2. These results are closer to what we would have predicted from the basic problem set than were the results with the Tileworld. In particular, LCFR-DSep does very well, generating much smaller search spaces than LCFR. However, it does slightly worse than ZLIFO. Recall that we saw the same pattern of performance on a subset of the basic problems, specifically on the Get-Paid/Uget-Paid problems. There, LCFR-DSep again improved on LCFR, but did not generate as small search spaces as ZLIFO. It turns out that there are similar factors influencing both sets of problems, and it is instructive to consider in some detail the planning done by ZLIFO and LCFR-DSep for the Get-Paid/Uget-Paid problems to understand what is occurring.

Like the Trains domain problems, the Get-Paid/Uget-Paid problems involve moving particular objects to specified locations. In the Get-Paid/Uget-Paid domain there are three objects: a paycheck, a dictionary, and a briefcase. As generally formulated, in the initial state all three are at home, and the paycheck is in the briefcase. The goal is to deposit the paycheck at the bank, bring the dictionary to work, and have the briefcase at home. Both the dictionary and the paycheck can be moved only in the briefcase. For a human, the solution to this problem is obvious. The dictionary must be put into the briefcase, and it must then be carried to work, where the dictionary is taken out. The briefcase must then be carried home. In addition, a stop must be made at the bank, either on the way to work or on the way home, at which point the paycheck must be taken out of the briefcase and deposited.

ZLIFO and LCFR-DSep take different paths in solving this problem. ZLIFO begins by forming plans to get the paycheck to the bank and the dictionary to work. These goals are selected first because they are forced: there is only one way to get the paycheck to the bank (carry it there), and similarly only one way to get the dictionary to the office (carry it there). In contrast, there are two possible ways to get the briefcase home: either by leaving it there (i.e., reusing the initial state) or by carrying it there from somewhere else (i.e., adding a new step). The LIFO mechanism then proceeds to complete the plans for achieving the goals of getting the paycheck to the bank and the dictionary to work, before beginning to work on the remaining goal, of getting the briefcase home. At this point, that goal is easy to solve. All that is needed is to plan a route home from wherever the briefcase is at the end of these two errands.

LCFR-DSep, like ZLIFO, begins by selecting the forced goals of getting the dictionary to the office and getting the paycheck to the bank. However, instead of next completing the plans for these goals, LCFR-DSep continues to greedily select least-cost flaws, and thus begins to work on achieving the goal of getting the briefcase home. Unfortunately, at this point it is not clear where the briefcase needs to be moved home *from*, and hence LCFR-DSep begins to engage in a lengthy process of "guessing" where the briefcase will be at the end of the other tasks, before it has planned for those tasks.[10]

---

10. The difficulty that LCFR-DSep encounters by greedily picking low-cost flaws might be reduced by doing a lookahead of several planning steps, to determine a more accurate repair cost. This is the approach taken in the branch-n mechanism in O-Plan (Currie & Tate, 1991). Significant overhead can be involved in such a strategy, however.

The key decision for the Get-Paid/Uget-Paid domain—and, as it turns out, for the Trains domain—is related to, but subtly different from the key decision in the Tileworld domain. For Get-Paid/Uget-Paid and Trains, the key insight for the planner is again that there is an important temporal ordering between goals. The goal of getting the briefcase home is going to have to be achieved after the goal of taking the dictionary to work. However, recognition of this constraint is not affected by separable-threat delay, as it was for the Tileworld. Instead, what happens in these domains is that a higher-cost flaw interacts with a lower-cost one, causing the latter to become fully constrained.

It is tempting to think that here finally is a case in which a LIFO-based strategy is advantageous. After all, for this example, by completely determining what you will do to achieve one goal, you make it much easier to know how to solve the another goal. But the use of ZLIFO (or an alternative LIFO-based strategy) does not guarantee that the interactions between high- and lower-cost flaws will be exploited. In particular if the interactions are among two or more unforced flaws, then the order of the goals in the agenda can lead ZLIFO to make an inefficient choice. Thus, when we modified the problem so that the briefcase was at work in the initial state, ZLIFO and LCFR-DSep both solved the problem very quickly (178 nodes for ZLIFO and 157 for LCFR-DSep). Note that this modification removes the problematic interaction between a low-cost and a high-cost flaw.

Finally, note that the effectiveness of the LIFO strategies is again heavily dependent on the the order in which preconditions are entered onto the open list. Figure 20 gives the node counts for the Trains domain with reverse precondition insertion. We once again plot only the strategies that can solve both Trains 1 and Trains2. In this case, there are only two such strategies: LCFR-DSep and DSep-LC. The strategies that rely on LIFO for open-condition selection, ZLIFO, DSep, DUnf-Gen, and UCPOP, all do significantly worse than they did when the preconditions were in the correct order. To the extent that LIFO helps in such domains, it appears to be because of its ability to exploit the decisions made by the system designers in writing the domain operators, as suggested by Williamson and Hanks (1996).

## 4.6 Computation Time

We have now covered the key questions we set out to address: what are the relative effects of alternative search-control strategies on search-space size, and, in particular, how can we reconcile the apparently conflicting approaches of LCFR and ZLIFO? We concluded that LCFR-DSep combines the main advantages for reducing search-space size of these two strategies, namely LCFR's use of a least-cost selection mechanism, at least for forced flaws, with ZLIFO's use of separable-threat delay. A final question concerns the price one has to pay to use LCFR-DSep—or for that matter, any of the alternative strategies. To achieve a reduction in search-space size, is it necessary to spend vastly more time in processing? Or do these strategies pay for themselves?

To answer these questions, we collected timing data on all our experiments. Figures 21 and 22 gives this data for the basic problems, for both the experiments run with the node limit and those run with the time limit. (As detailed in Appendix A, the results for the experiments with the node limit and the time limit were very similar.) Because we saw little influence of precondition ordering on the basic problems, we analyze only the data for the default precondition ordering. We show one graph with all the strategies, and another that includes only the "leading strategies", to make it possible to see the distinctions among them.

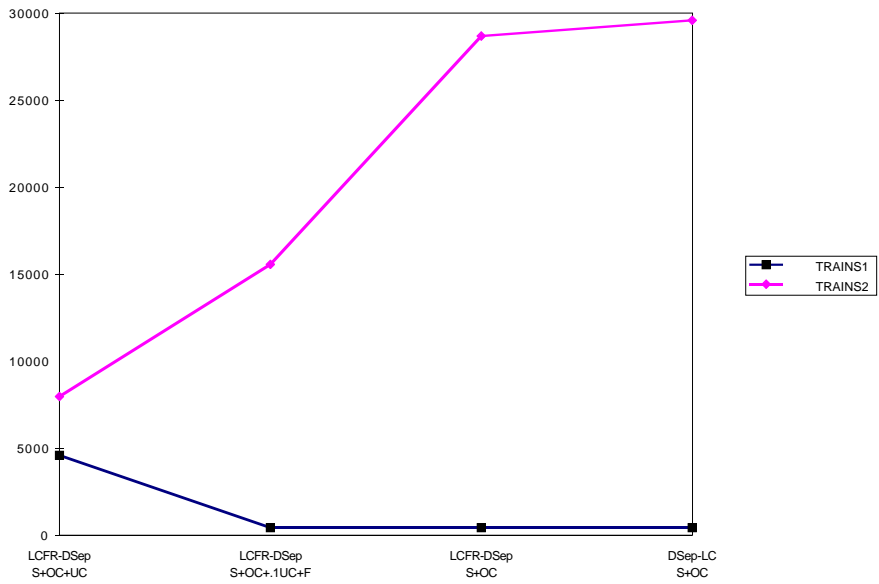

Figure 20: Trains Problems: Node Counts with Reversed Precondition Insertion

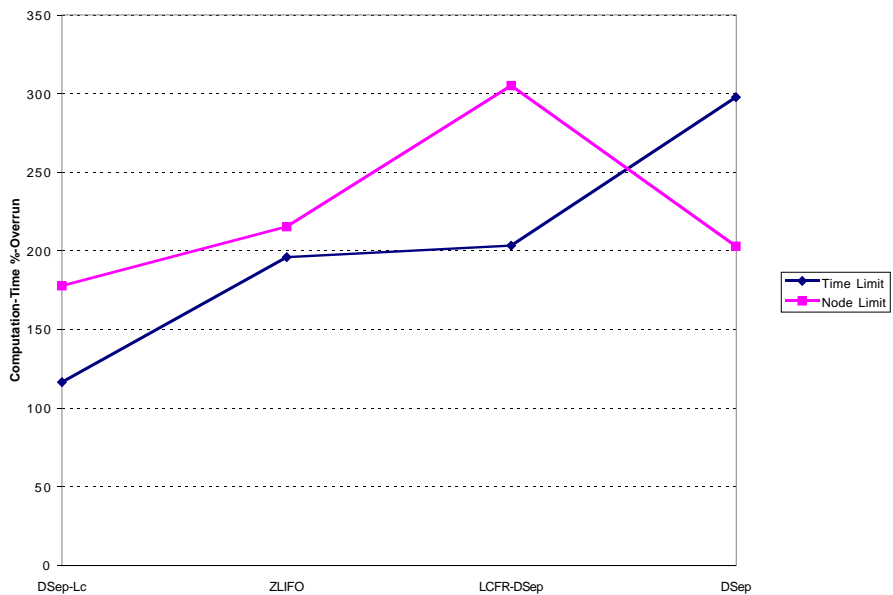

Figure 21: Basic Problems: Aggregate Computation Time Performance for Leading Strategies

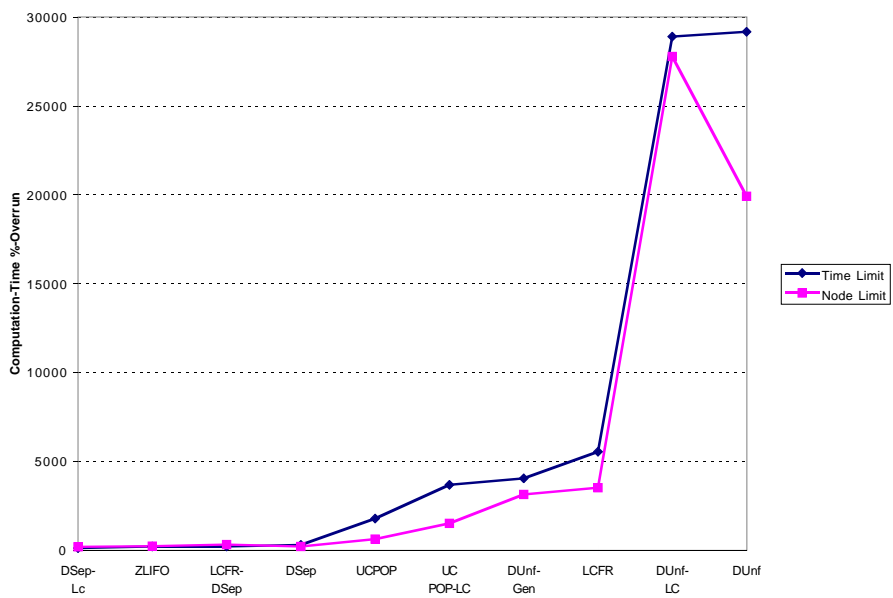

Figure 22: Basic Problems: Aggregate Computation Time Performance

The timing data show that LCFR-DSep does, by and large, pay for its own overhead on the basic problems by generating smaller search spaces (and therefore having to process fewer nodes). When run with a time limit, LCFR-DSep's time performance is almost identical with ZLIFO's, despite the fact that repair cost computations are more expensive than the stack-popping of a LIFO strategy. When run with a node limit, LCFR-DSep does show worse time performance than ZLIFO in aggregate, but still performs markedly better than most of the other strategies. The change in relative performance results from the cases in which both strategies fail at the node limit: LCFR-DSep takes longer to generate 10,000 nodes.

Another interesting observation is that DSep-LC has the best time performance of all on the basic problem set. This should perhaps not be a surprise, because DSep-LC closely approximates LCFR-DSep. It differs primarily in its preference for nonseparable threats, which in any case will tend to have low repair costs. Whenever a node includes a nonseparable threat, DSep-LC can quickly select that threat, without having to compute repair costs. This speed advantage outweighs the cost of processing the extra nodes it sometimes generates.

Figures 23–26 provide the timing data for the Trains and Tileworld domains.[11] Here there are no real surprises. The computation times taken parallel quite closely the size of the search spaces generated. The strategies that generate the smallest search spaces are also the fastest. With the Trains problems, we again see the DSep-LC can serve as

---

11. We have omitted the strategies that did very poorly, performing worse *both* on the node- and time-limit experiments than did any of the strategies graphed. Note that we ran the reverse-order experiments only with a node limit.

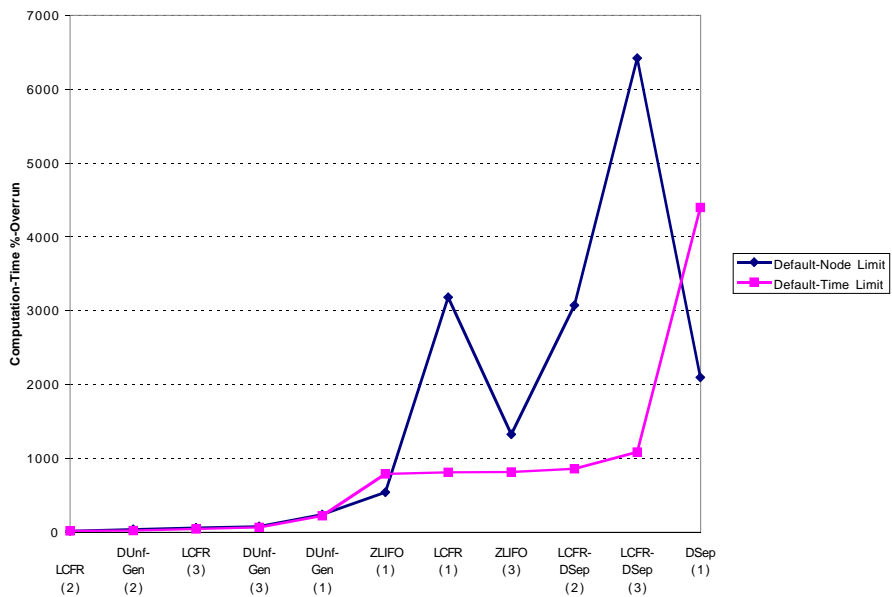

Figure 23: Tileworld Problems: Aggregate Computation Time Performance for Leading Strategies

a good approximation technique for LCFR-DSep. Although it generates more nodes than LCFR-DSep, it is somewhat faster.[12]

## 5. Conclusion

In this paper, we have synthesized much of the previous work on flaw selection for partial-order causal link planning, showing how earlier studies relate to one another, and have developed a concise notation for describing alternative flaw selection strategies.

We also presented the results of a series of experiments aimed at clarifying the effects of alternative search-control preferences on search-space size. In particular, we aimed at explaining the comparative performance of the LCFR and ZLIFO strategies. We showed that neither of these flaw selection strategies consistently generates smaller search spaces, but that by combining LCFR's least-cost approach with the delay of separable threats that is included in the ZLIFO strategy, we obtain a strategy—LCFR-DSep—whose space performance was nearly always as good as the better of LCFR or ZLIFO on a given problem. We therefore concluded that much of ZLIFO's advantage relative to LCFR is due to its delay of separable threats rather than to its use of a LIFO strategy. Although we were unable to resolve the question of whether least-cost selection is required for unforced, as well as forced flaws, we found no evidence that a LIFO strategy for unforced flaws was better. On the other hand, separable-threat delay is clearly advantageous. An open question is exactly why it is so advantageous. We have conducted preliminary experiments that suggest that much of the search-space reduction that results from delaying separable threats can also be achieved by making separation systematic, something that UCPOP v.4 does not do.

---

12. In interpreting the Trains timing data, it is important to note that some of the strategies shown—notably UCPOP, UCPOP-LC, and Dunf, failed to solve Trains2 within either the node or the time limit.

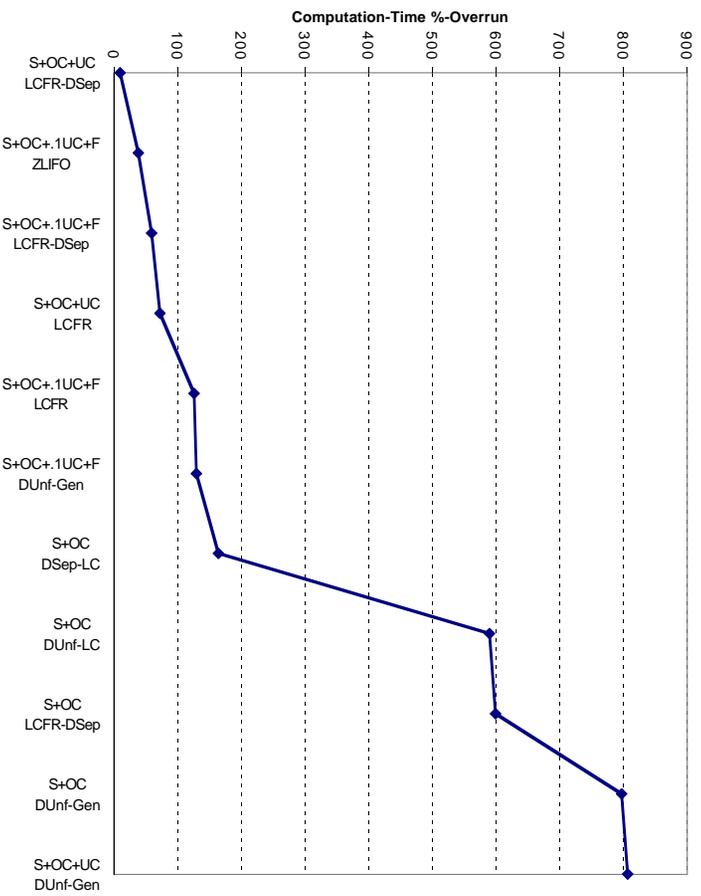

Figure 24: Tileworld Problems: Aggregate Computation Time Performance for Leading Strategies with Reversed Precondition Insertion

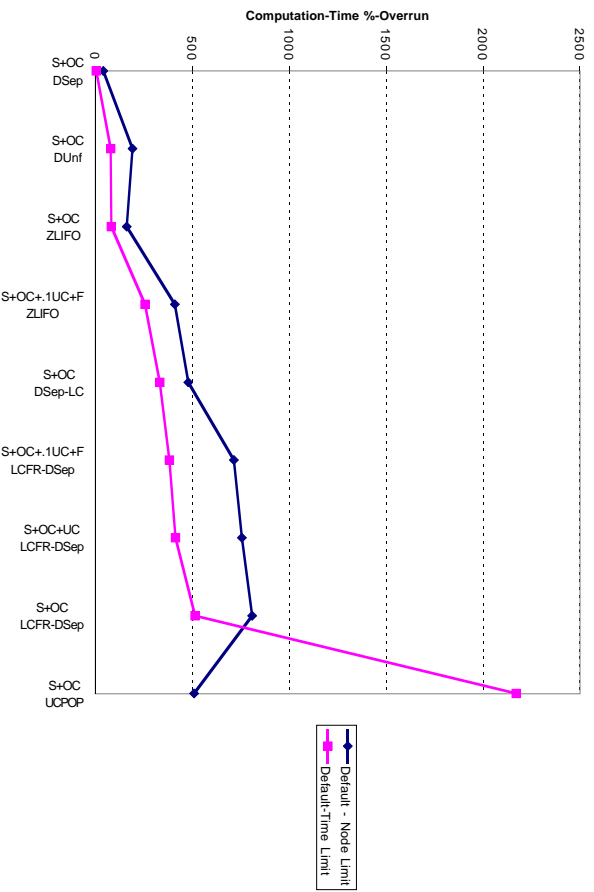

Figure 25: Trains Problems: Aggregate Computation Time Performance for Leading Strategies

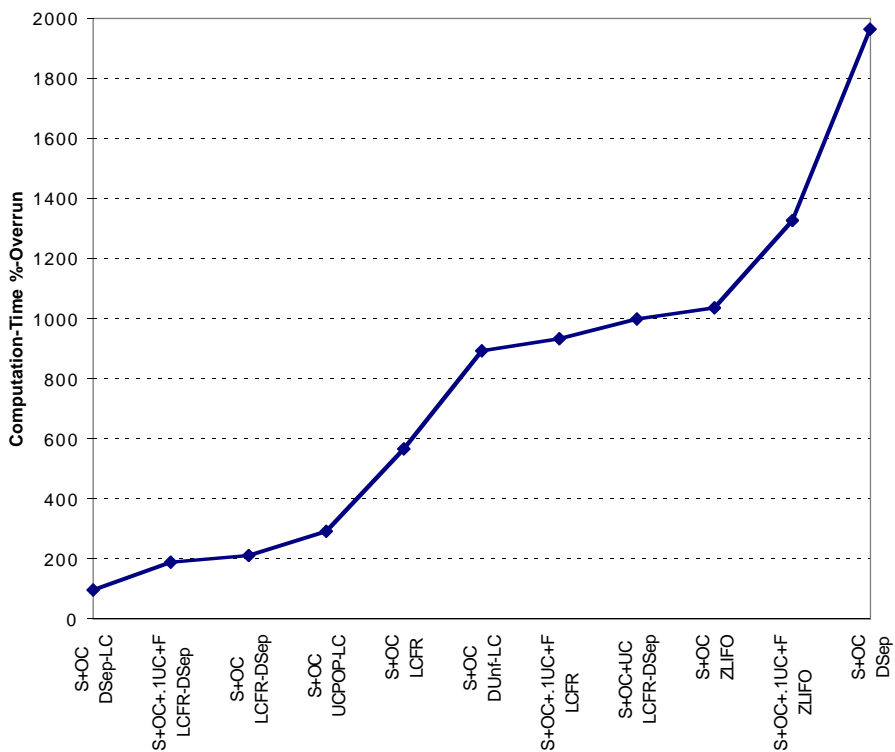

Figure 26: Trains Problems: Aggregate Computation Time Performance for Leading Strategies with Reversed Precondition Insertion

We also considered the question of computation time, and showed that often LCFR-DSep only requires computation time comparable to that of ZLIFO. LCFR-DSep can therefore be seen as paying for its own computational overhead by its search-space reduction. Moreover, Peot and Smith's DSep-LC provides a good approximation of LCFR-DSep: although it produces somewhat larger search spaces, it does so more quickly.

These conclusions, however, are tempered by the fact that for certain clusters of problems, our combined strategy, LCFR-DSep, does not generate minimal search spaces. As we saw, for the Tileworld problems, what is most important is to recognize the need for a particular temporal ordering among plan steps, and this recognition can be obtained by resolving separable threats early. For the Trains and Get-Paid/Uget-Paid domains, what matters most is recognizing that a particular effect can in fact only be achieved in one way, and this is only recognized when a particular flaw is selected—a flaw which happens generally not to be the least cost flaw available. The lesson to be learned from these sets of problems is that although we now understand the reasons that LCFR and ZLIFO perform the way they do, and how to combine the best features of both to create good default strategies for POCL planning, it is clear that domain-dependent characteristics such as those we identified in the Trains and Tileworld domains must still be taken into account in settling on a flaw selection strategy for any domain.

## Acknowledgments

Martha Pollack's work on this project has been supported by the Air Force Office of Scientific Research (F49620-96-1-0403) and an NSF Young Investigator's Award (IRI-9258392). David Joslin has been supported by Rome Labs (RL)ARPA (F30602-95-1-0023) and an NSF CISE Postdoctoral Research award (CDA-9625755). Massimo Paolucci has been supported by the Office of Naval Research, Cognitive and Neural Sciences Division (N00014-91-J-1694).

We are very grateful to Alfonso Gerevini for providing us with the code he used in his earlier study, and allowing us to use it in our experiments. We would also like to thank Arthur Nunes and Yazmine DeLeon, who assisted us in carrying out experiments done in the preliminary stages of this work. Finally, we thank Alfonso Gerevini, Len Schubert, Michael Wellman, and the anonymous reviewers for their helpful comments on this work.

## Appendix A: Ruling Out Ceiling Effects

For the data collected using a node limit, we examined all the problems in which at least one of the strategies hit the node limit. Table 27 gives the *second worst* node count for all such problems. It shows that, for all the basic problems in which at least one strategy failed, and at least one other succeeded, the second-worst strategy generally created fewer than 7000 nodes.

Similarly, for the Trains and Tileworld problems, in all such cases except TW3, the second-worst strategy took fewer than 50,000 nodes (and in TW3 it took 89,790). Recall that the node limit for the basic problems was 10,000 nodes, while for the Trains and Tileworld problems it was 100,000 nodes. It is thus clear that the strategies that hit the node limit are doing substantially worse than the strategies that succeed. Even if they were to succeed by increasing the node limit slightly, their comparative performance would still be poor.

| PROBLEM | Default | Reverse |
|---|---|---|
| HANOI | 2919 | 2952 |
| R-TEST2 | 7567 | 5227 |
| MONKEY-TEST2 | 3744 | 5200 |
| MONKEY-TEST3 | 10000 | 10000 |
| GET-PAID2 | 129 | 129 |
| GET-PAID3 | 6431 | 6431 |
| GET-PAID4 | 1625 | 1625 |
| FIXIT | 10000 | 10000 |
| HO-DEMO | | 10000 |
| FIXB | 10000 | 3184 |
| UGET-PAID2 | 175 | 175 |
| UGET-PAID3 | 4725 | 4725 |
| UGET-PAID4 | 2894 | 2894 |
| PRODIGY-P22 | 8265 | 9264 |
| MOVE-BOXES | 4402 | 2687 |
| MOVE-BOXES-1 | 10000 | 10000 |
| | | |
| TRAINS2 | 22351 | 29585 |
| TRAINS3 | 100000 | 100000 |
| | | |
| TW-2 | | 11620 |
| TW-3 | 89790 | 401 |
| TW-4 | 3844 | 1266 |
| TW-5 | 49024 | 20345 |
| TW-6 | 1722 | 3040 |

Figure 27: Second-Worst Node Counts on Problems with Failing Strategies

Thus, by using the node limits we imposed, we are not making any strategies look worse than they actually are. On the other hand, in computing %-overrun, we may be making some strategies look better than they actually are, because we use a value of 10,000 (or 100,000) nodes generated when a strategy hits the limit, and the actual number of nodes it might take, if run to completion, could be significantly higher. This is why, in our analyses, we considered both the absolute performance of strategies on individual problems, and their aggregate performance, as measured by average %-overrun.

We also compared the experiments that were run with a time limit and those that were run with a node limit. For the basic problem set, the time limit of 100 seconds was high enough that, in most cases, strategies could compute significantly more nodes than they could with the node cutoff. Nonetheless, the results were almost identical. In nearly all cases, if a strategy failed with the node cutoff, it also failed with the time limit cutoff. There were only four exceptions to this:

1. Hanoi: With the 10,000 nodes limit, DSep fails, while with the 100 second time limit, it succeeds, taking 46,946 nodes.

2. Uget-Paid3: With the 10,000 node limit, UCPOP-LC fails, while with the 100 second time limit, it succeeds, taking 37,951 nodes.

3. Uget-Paid4: With the 10,000 node limit, UCPOP-LC fails, while with 100 second time limit, it succeeds, taking 23,885 nodes.

4. Fixit: With the 10,000 nodes limit, DSep-LC, UCPOP-LC, and ZLIFO fail, while with the 100 second time limit, they succeed in 12,732, 13,510, and 20,301 nodes respectively. All the other strategies fail to solve this problem under either limit.

There was a similarly strong correspondence between the results we obtained on the Trains and Tileworld problems using a node limit and a time limit. In a few cases, a strategy that was able to succeed within the 100,000 node limit was *not* able to succeed within the 1,000 second time limit. The nature of these problems is that the computation time per node can be very great. Specifically,

1. On TW3, DUnf succeeded in 56,296 nodes when run with a node limit, but failed with the 1,000 second time limit.

2. On TW4, LCFR-DSep (with an S+OC node-selection strategy) succeeded in 69,843 nodes, but failed on the time limit.

3. On TW5, LCFR-DSep (with an S+OC+UC node-selection strategy) succeeded in 49,024, but failed on the limit.

4. On TW6, LCFR (with an S+OC node-selection strategy) succeeded in 4,506 nodes, but failed with the time limit.

In only one case did a strategy fail under the node limit but succeed within the time limit:

1. On TW3, DSep (with an S+OC node-selection strategy) failed with a 100,000 node limit, but succeeded with 134,951 nodes using a 100 second time limit. Note that this is significantly worse than the second worst strategy, which solved this problem generating 89,790 nodes.

Given this close correspondence between the experiments with node and time limits, we collected only node-limit data for the experiments in which we reversed the precondition insertion.

# References

Allen, J. F., Schubert, L. K., Ferguson, G. M., Heeman, P. A., Hwant, C. H., Kato, T., Light, M., Margin, N. G., Miller, B. W., Poesio, M., & Traum, B. R. (1995). The TRAINS project: A case study in building a conversational planning agent. *Experimental and Theoretical Artificial Intelligence*, *7*, 7–48.

Chapman, D. (1987). Planning for conjunctive goals. *Artificial Intelligence*, *32*(3), 333–378.

Currie, K., & Tate, A. (1991). O-plan: The open planning architecture. *Artificial Intelligence*, *52*, 49–86.

Etzioni, O., Hanks, S., Weld, D., Draper, D., Lesh, N., & Williamson, M. (1992). An approach to planning with incomplete information. In *Proceedings of the Third International Conference on Principles of Knowledge Representation and Reasoning*, pp. 115–125.

Gerevini, A. (1997). Personal communication.

Gerevini, A., & Schubert, L. (1996). Accelerating partial-order planners: Some techniques for effective search control and pruning. *Journal of Artificial Intelligence Research*, *5*, 95–137.

Joslin, D. (1996). *Passive and Active Decision Postponement in Plan Generation*. Ph.D. thesis, Intelligent Systems Program, University of Pittsburgh.

Joslin, D., & Pollack, M. E. (1994). Least-cost flaw repair: A plan refinement strategy for partial-order planning. In *Proceedings of the Twelfth National Conference on Artificial Intelligence (AAAI)*, pp. 1004–1009 Seattle, WA.

Joslin, D., & Pollack, M. E. (1996). Is "early commitment" in plan generation ever a good idea?. In *Proceedings of the Thirteenth National Conference on Artificial Intelligence (AAAI)*, pp. 1188–1193 Portland, OR.

Kambhampati, S., Knoblock, C. A., & Yang, Q. (1995). Planning as refinement search: A unified framework for evaluating design tradeoffs in partial-order planning. *Artificial Intelligence*, *76*(1-2), 167–238.

Kumar, V. (1992). Algorithms for constraint-satisfaction problems: A survey. *AI Magazine*, *13*(1), 32–44.

McAllester, D., & Rosenblitt, D. (1991). Systematic nonlinear planning. In *Proceedings of the Ninth National Conference on Artificial Intelligence*, pp. 634–639 Anaheim, CA.

Pednault, E. P. D. (1988). Synthesizing plans that contain actions with context-dependent effects. *Computational Intelligence*, *4*(4), 356–372.

Penberthy, J. S., & Weld, D. (1992). UCPOP: A sound, complete, partial order planner for ADL. In *Proceedings of the Third International Conference on Knowledge Representation and Reasoning*, pp. 103–114 Cambridge, MA.

Peot, M., & Smith, D. E. (1992). Conditional nonlinear planning. In *Proceedings of the First International Conference on AI Planning Systems (AIPS-92)*, pp. 189–197 College Park, MD.

Peot, M., & Smith, D. E. (1993). Threat-removal strategies for partial-order planning. In *Proceedings of the Eleventh National Conference on Artificial Intelligence*, pp. 492–499 Washington, D.C.

Pollack, M. E., & Ringuette, M. (1990). Introducing the Tileworld: Experimentally evaluating agent architectures. In *Proceedings of the Eighth National Conference on Artificial Intelligence*, pp. 183–189 Boston, MA.

Russell, S., & Norvig, P. (1995). *Artificial Intelligence: A Modern Approach*. Prentice Hall, Englewood Cliffs, NJ.

Russell, S. J. (1992). Efficient memory-bounded search algorithms. In *Proceedings of the Tenth European Conference on Artificial Intelligence*, pp. 1–5.

Smith, D. E., & Peot, M. A. (1994). A note on the DMIN strategy. Unpublished manuscript.

Srinivasan, R., & Howe, A. E. (1995). Comparison of methods for improving search efficiency in a partial-order planner. In *Proceedings of the 14th International Joint Conference on Artificial Intelligence*, pp. 1620–1626.

Tate, A., Drabble, B., & Dalton, J. (1994). Reasoning with constraints within O-plan2. Tech. rep. ARPA-RL/O-Plan2/TP/6 V. 1, AIAI, Edinburgh.

Tsang, E. (1993). *Foundations of Constraint Satisfaction*. Academic Press.

Tsuneto, R., Erol, K., Hendler, J., & Nau, D. (1996). Commitment strategies in hierarchical task network planning. In *Proceedings of the Thirteenth National Conference on Artificial Intelligence (AAAI)*, pp. 526–542 Portland, OR.

Weld, D. S. (1994). An introduction to least commitment planning. *AI Magazine*, *15*(4), 27–61.

Wilkins, D. E. (1988). *Practical Planning: Extending the Classical AI Paradigm*. Morgan Kaufmann, San Mateo, CA.

Wilkins, D. E., & Desimone, R. V. (1994). Applying an AI planner to military operations planning. In Fox, M., & Zweben, M. (Eds.), *Intelligent Scheduling*, pp. 685–708. Morgan Kaufmann Publishers, San Mateo, CA.

Williamson, M., & Hanks, S. (1996). Flaw selection strategies for value-directed planning. In *Proceedings of the Third International Conference on Artificial Intelligence Planning Systems*, pp. 237–244.